\pdfoutput=1
\PassOptionsToPackage{most,breakable}{tcolorbox}
\PassOptionsToPackage{dvipsnames,table}{xcolor}
\PassOptionsToPackage{numbers, compress}{natbib}
\documentclass[11pt]{article}
\usepackage[final]{ropedia}
\usepackage[T1]{fontenc}
\usepackage[utf8]{inputenc}
\usepackage{amsmath,amssymb,amsfonts}
\usepackage{mathpazo}
\usepackage[scaled]{helvet}
\usepackage{microtype}

\usepackage{amsmath,amsfonts,bm}

\def\eqref#1{equation~\ref{#1}}

\def\1{\bm{1}}

\DeclareMathAlphabet{\mathsfit}{\encodingdefault}{\sfdefault}{m}{sl}
\SetMathAlphabet{\mathsfit}{bold}{\encodingdefault}{\sfdefault}{bx}{n}

\usepackage{url}
\usepackage{booktabs}
\usepackage{array}
\usepackage{multirow}
\usepackage{graphicx}
\usepackage{enumitem}
\usepackage{float}
\usepackage{tikz}
\usetikzlibrary{arrows.meta, positioning, fit, calc}
\usepackage{xspace}
\usepackage{listings}

\newcommand{\sysname}{S-Agent 2.0\xspace}
\newcommand{\code}[1]{{\small\ttfamily #1}}
\definecolor{upgreen}{RGB}{20,130,60}
\definecolor{dnred}{RGB}{200,40,40}
\newcommand{\up}[1]{\,{\scriptsize\textcolor{upgreen}{(+#1)}}}
\newcommand{\dn}[1]{\,{\scriptsize\textcolor{dnred}{($-$#1)}}}
\lstdefinestyle{sccode}{basicstyle=\ttfamily\tiny,breaklines=true,breakatwhitespace=false,columns=fullflexible,keepspaces=true,frame=single,framerule=0.3pt,rulecolor=\color{black!35},backgroundcolor=\color{black!3},xleftmargin=2pt,framexleftmargin=2pt,aboveskip=2pt,belowskip=2pt,lineskip=-1pt}
\lstdefinestyle{scout}{basicstyle=\ttfamily\tiny\color{black!80},breaklines=true,breakatwhitespace=false,columns=fullflexible,keepspaces=true,frame=leftline,framerule=1.2pt,rulecolor=\color{black!45},xleftmargin=4pt,framexleftmargin=2pt,aboveskip=1pt,belowskip=3pt,lineskip=-1pt}

\firstpagelogos{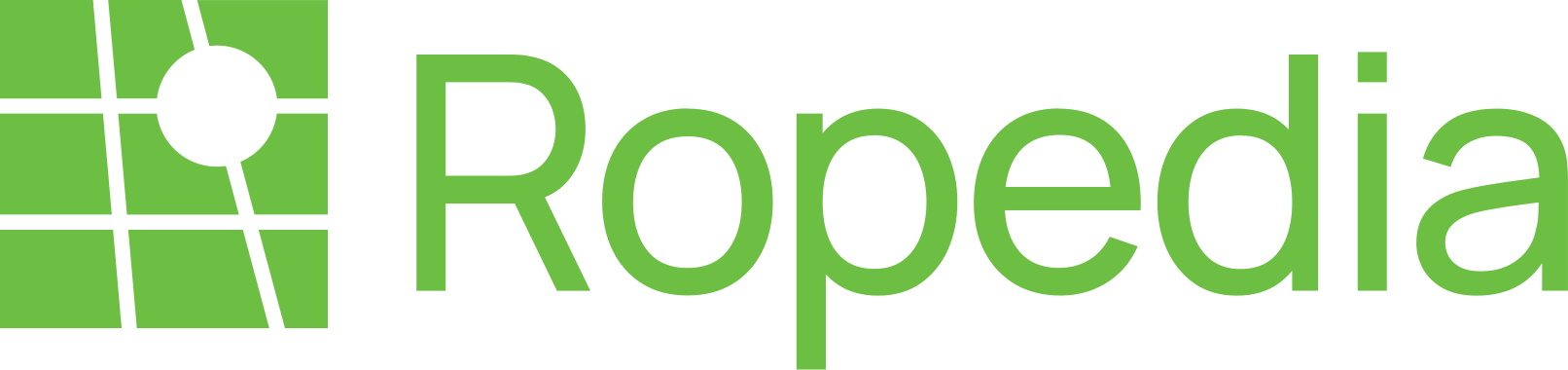}{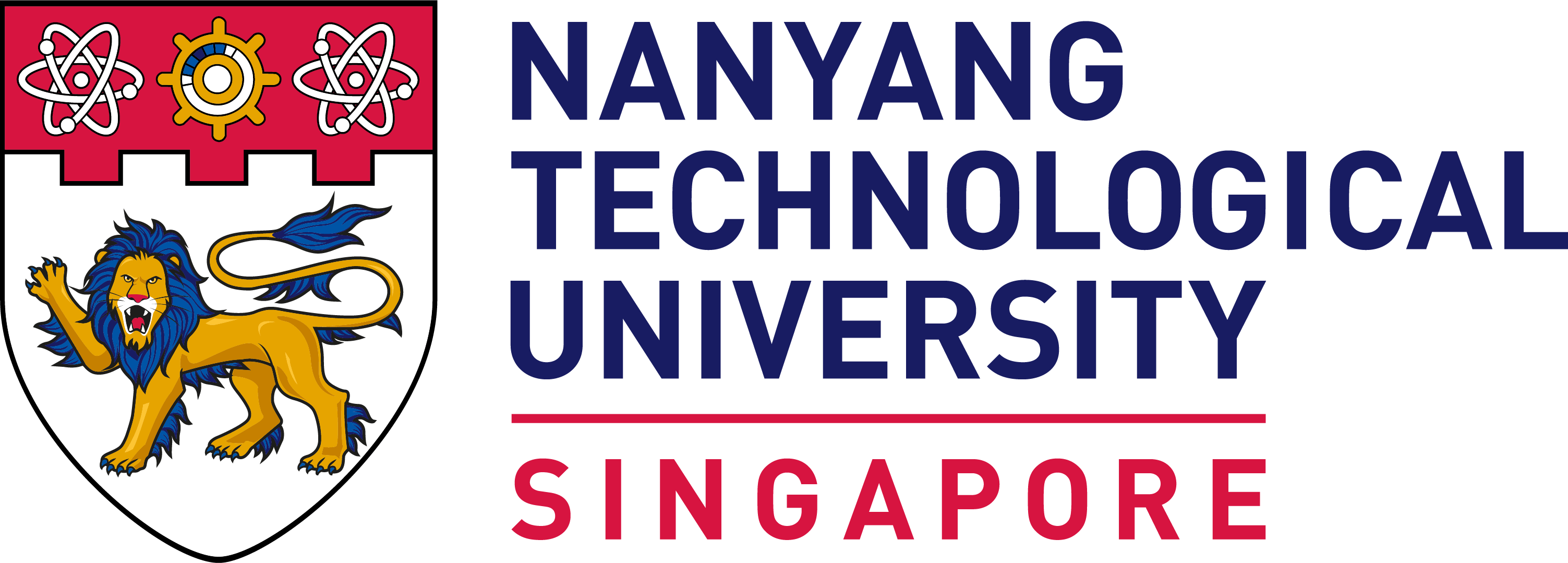}{}
\abstractlogo{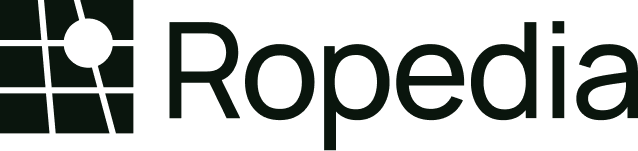}
\title{\includegraphics[height=30pt]{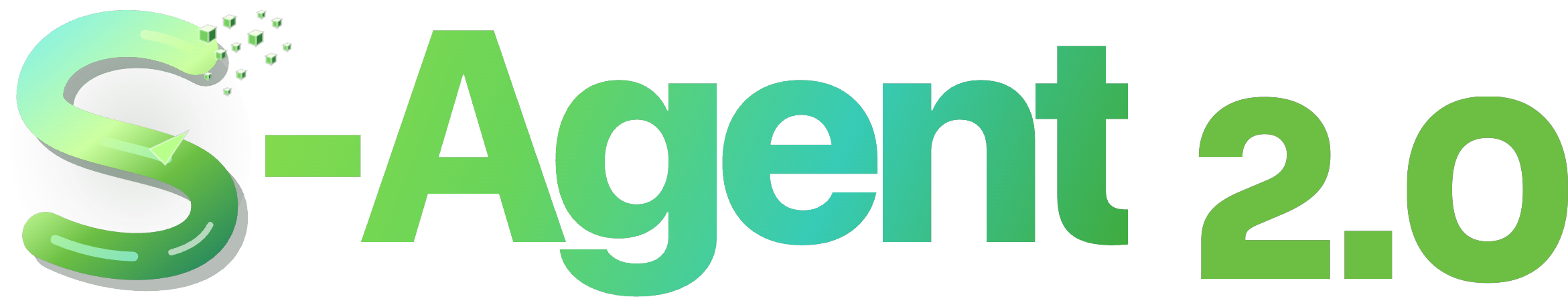}}
\subtitle{Spatial Reasoning with a General Coding Agent}

\newcommand{\coremark}{\ensuremath{^{*}}}
\newcommand{\corrmark}{\ensuremath{^{\dagger}}}
\newcommand{\authorfootnotes}{%
  \begingroup
  \renewcommand{\thefootnote}{}%
  \footnotetext{$*$ Equal contributors. $\dagger$ Corresponding author.}%
  \endgroup
  \addtocounter{footnote}{-1}%
}
\author{
  \AuthorName{Rui Gong}{1\coremark}\quad
  \AuthorName{Ruoyu Wang}{1\coremark}\quad
  \AuthorName{Yalun Dai}{1\coremark}\quad
  \RopediaAuthorName{Hao Li}{1}\\[0.55ex]
  \AuthorName{Sidi Yang}{2}\quad
  \AuthorName{Qihao Zhao}{1}\quad
  \AuthorName{Hongyu Li}{1}\quad
  \AuthorName{Zeyang Guan}{1}\\[0.55ex]
  \RopediaAuthorName{Fangzhou Hong}{1}\quad
  \RopediaAuthorName{Zhaoxi Chen}{1}\quad
  \AuthorName{Kim-Hui Yap}{1\corrmark}\quad
  \RopediaAuthorName{Ziwei Liu}{1}\\[1.0ex]
  {\normalfont\footnotesize
  \mbox{\textsuperscript{1}Nanyang Technological University}\quad
  \mbox{\textsuperscript{2}University of Hong Kong}\quad
  \mbox{\textsuperscript{\ropediamark}Ropedia}\quad
  \paperresourceicon{\resourcegithubicon}{Code}{https://ropedia.github.io/S-Agent-2.0}{ropedia/S-Agent-2.0}}
}

\begin{document}
\maketitle
\authorfootnotes

\begin{figure}[H]
\centering
\includegraphics[width=\linewidth]{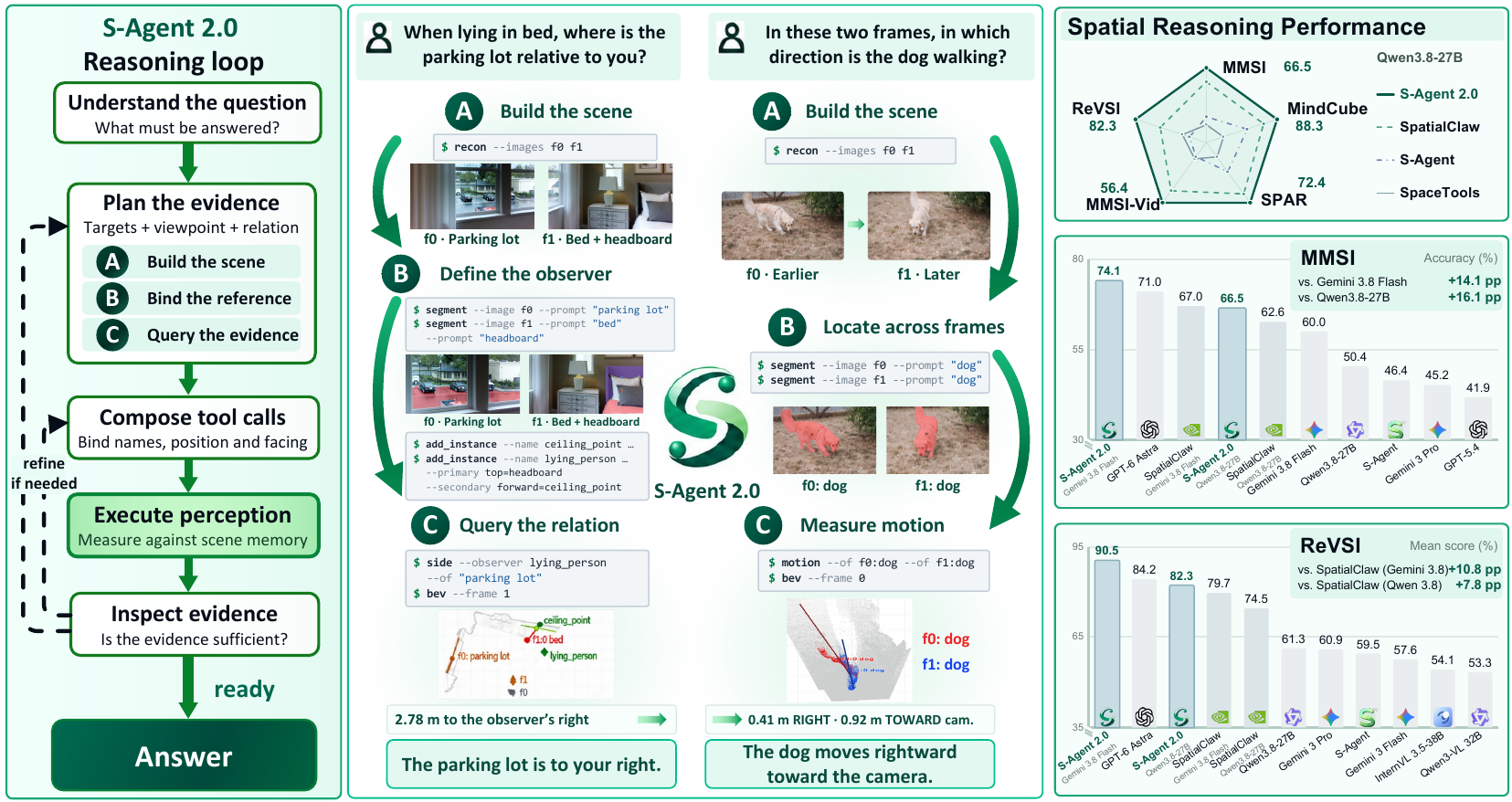}
\caption{\sysname{} is a spatial reasoning agent built on a general
coding agent. A command-line perception toolkit lets the agent close the
evidence loop, acquiring scene evidence, remembering it in a persistent
scene memory, and interpreting every reading in plain spatial language.
\sysname{} delivers strong performance: it leads MMSI-Bench and ReVSI and
stays ahead of the specialized agentic system SpatialClaw on every
benchmark and backbone we test.}
\label{fig:teaser}
\end{figure}

\begin{abstract}
Spatial reasoning asks a model to work out where objects are in 3D space,
how they relate to one another, and how they move. Agentic VLMs have
recently brought large improvements on such tasks by letting the model
actively gather evidence with 3D perception tools instead of answering in a
single pass. Even so, on many questions the tools return the right
evidence and the model still does not arrive at the correct answer: the
gap between model and tools, where raw readings become usable
conclusions, is left unfilled. We present \sysname{}, a spatial
reasoning agent that equips an off-the-shelf \emph{general coding
agent} with a command-line perception toolkit, and we design the
interface around one principle: help the VLM \emph{close the evidence
loop}---\emph{acquire} scene evidence, \emph{remember} it in a
persistent scene memory, and \emph{interpret} each reading in plain
spatial language. This design keeps the full flexibility of a coding
agent, and every reading it acts on stays grounded in the tools'
geometry. On MMSI-Bench and ReVSI, the toolkit improves a strong
open-source model, Qwen3.8-27B, by $+13.9$ and $+20.5$ points over the
same agent without tools, and a strong closed-source model,
Gemini-3.8-Flash, by $+11.2$ and $+9.6$ points.
With the open backbone \sysname{} reaches $66.5$ on MMSI-Bench and $82.3$
on ReVSI, ahead of the specialized system SpatialClaw on the same backbone
($62.6$ and $74.5$), and with Gemini-3.8-Flash it reaches $74.1$ and
$90.5$ against its $67.0$ and $79.7$. Both tool gains shrink as the
backbone grows more capable, but SpatialClaw's shrinks far more on ReVSI,
from $+13.2$ to $+2.7$, while ours stays at $+9.6$.
\sysname{} also works across different general coding agent harnesses,
such as Codex and Claude Code.
\end{abstract}

\section{Introduction}
\label{sec:intro}

Spatial intelligence, the ability to understand where things are, how they
relate in three dimensions, and how the world would look as one moves
through it, is foundational to visual
perception~\citep{chen2024spatialvlm,vsi_bench} and essential for any
agent that operates in the physical
world~\citep{driess2023palme,brohan2023rt2,song2025robospatial,geiger2012kitti,caesar2020nuscenes}.
Agentic VLMs such as S-Agent~\citep{sagent} have recently brought large improvements here: instead of
answering in a single pass, the model actively gathers evidence with 3D
perception tools while it reasons~\citep{luo2026pySpatial,chen2026spacetools}. More recent work
such as SpatialClaw~\citep{spatialclaw} responds with a specialized
harness: a custom state machine that drives the VLM through a hand-designed
loop (plan, act, reflect) over a persistent code kernel with a library of
3D perception and geometry utilities, leveraging the flexibility of code
to bridge the gap. Even so, on many
questions the tools return the right evidence and the model still does not
arrive at the correct answer (see Figure~\ref{fig:gap}).

Watching such a system work reveals where it actually fails. A spatial answer
is grounded only when a complete \textbf{evidence loop} closes:
\emph{acquire} the measurement, \emph{remember} it while the rest of the
evidence is gathered, and \emph{interpret} the reading into understanding.
Specialized systems invest almost entirely in
acquisition, in the form of ever richer perception and geometry libraries,
and leave the other two steps to the model. In our analysis of SpatialClaw
trajectories, acquisition is rarely what breaks. The failures come in two
forms. In the first, the model misreads a reading it holds: it flips the
sign of a rotation it measured correctly, or reads a correct number in the
wrong coordinate frame or the wrong sense. In the second, the evidence
never reaches the answer at all: the measurement sits in the workspace
while the model eyeballs the pictures, or a correct reading is discarded
for an unrelated stated reason. The second form exposes a break between
collecting evidence and drawing the conclusion. Once the readings are in,
the reasoning that produces the answer runs detached from them, and the
model falls back on its own perception of the scene. On the surface only
interpretation goes wrong; underneath, the evidence that was acquired is
never fully used. Closing the loop therefore places a requirement on each
step. Acquisition has to combine easily with the VLM's own perception, so
that what the model recognizes and what the tools measure enter the same
scene. Acquired evidence needs a memory that holds it without effort, so
that it stays within reach until the answer is written. And interpretation
needs spatial language, so that a reading arrives in the terms the
question is asked in. 

\subsection{S-Agent 2.0}
\label{sec:intro-sagent2}

\sysname{} is our answer to those requirements: a spatial reasoning agent
assembled from a stock general-purpose coding agent \citep{openai2025codex,anthropic2025claudecode,wang2025openhands} and a command-line
perception toolkit designed to meet them. General-purpose coding agents (the
harnesses behind modern software-engineering agents) already provide the
scaffolding a spatial reasoner needs (a shell, a file system, code
execution, long-horizon context management, a mature tool-calling loop),
and two of their habits bear directly on the requirements above. First,
looking at a frame or at a segmentation mask is an ordinary file view
in the agent's stream, so the model's own look at the scene and the
tools' readouts interleave in one conversation, with no separate image
channel between them.\citep{yang2024sweagent} Second, a coding agent inspects and revises by
reflex: it views an output, corrects what is wrong, and runs again.
\citep{yang2024sweagent,critic,shinn2023reflexion}
Applied to a scene, that reflex is what joins the two kinds of
perception, since the model's own perception is checked against what the
tools registered before any reading is taken. What these agents lack is
perception itself; we keep the harness stock and add it as commands. The
design effort goes into the toolkit, which meets the three requirements
one by one.

Concretely, (i)~acquisition combines two kinds of perception, the tools'
measurements and the model's own: the model can register what it
perceives or imagines as instances in the scene and drop what the tools
got wrong, so both kinds land on one geometric footing;
(ii)~everything measured persists in a scene memory that later
steps query by name instead of re-perceiving, so acquired evidence stays
within reach until the answer is written; and (iii)~every measurement
comes back as an answer-ready reading stated in plain spatial language
with its interpretation attached.

Our contributions can be summarized as follows:
\begin{itemize}[leftmargin=1.3em, itemsep=2pt, topsep=2pt]
  \item We build a spatial reasoning agent from a stock general-purpose
  coding agent and a command-line perception toolkit, 
  and show that the method works across different general
  coding agent harnesses (\S\ref{sec:exp-harness}).
  \item We design a command-line perception toolkit that closes the
  evidence loop for the agent: each call acquires scene evidence, writes it
  into a persistent scene memory, and returns its reading already
  interpreted in spatial language, and the model's own perception is
  grounded in the same scene (\S\ref{sec:method}).
  \item \sysname{} delivers strong performance. On MMSI-Bench and ReVSI,
  the perception toolkit adds $+13.9$ and $+20.5$ points over the same
  agent without tools, on the already strong Qwen3.8-27B backbone
  (\S\ref{sec:exp-main}), and \sysname{} surpasses the state-of-the-art
  agentic spatial reasoning method, whose tool gain on ReVSI shrinks far
  more on the stronger frontier model Gemini-3.8-Flash while ours
  remains substantial (\S\ref{sec:exp-model}).
\end{itemize}

\subsection{Takeaways}
\label{sec:takeaways}

\begin{enumerate}[leftmargin=1.6em, itemsep=2pt, topsep=2pt]
  \item \textbf{The method works across coding-agent harnesses.}
  \sysname{} works across coding-agent harnesses, including OpenHands~\citep{wang2025openhands}, Codex~\citep{openai2025codex} and Claude Code~\citep{anthropic2025claudecode}
  (\S\ref{sec:exp-harness}).
  \item \textbf{The coding agent's inspect-and-revise habit joins the
  model's perception with the tools'.} The model registers what it
  perceives or imagines into the scene memory, drops what the tools got
  wrong (\S\ref{sec:method-division}), and cross-checks the memory
  against the perception outputs (\S\ref{sec:exp-qualitative}).
  \item \textbf{The remaining gap is model-to-tool grounding.}
  The model's reasoning is mostly right; what remains is grounding the
  question into a measurement, which fails in two classes
  (\S\ref{sec:exp-qualitative}).
\end{enumerate}

\section{Method}
\label{sec:method}

\begin{figure}[t]
\begin{center}
\includegraphics[width=0.95\linewidth]{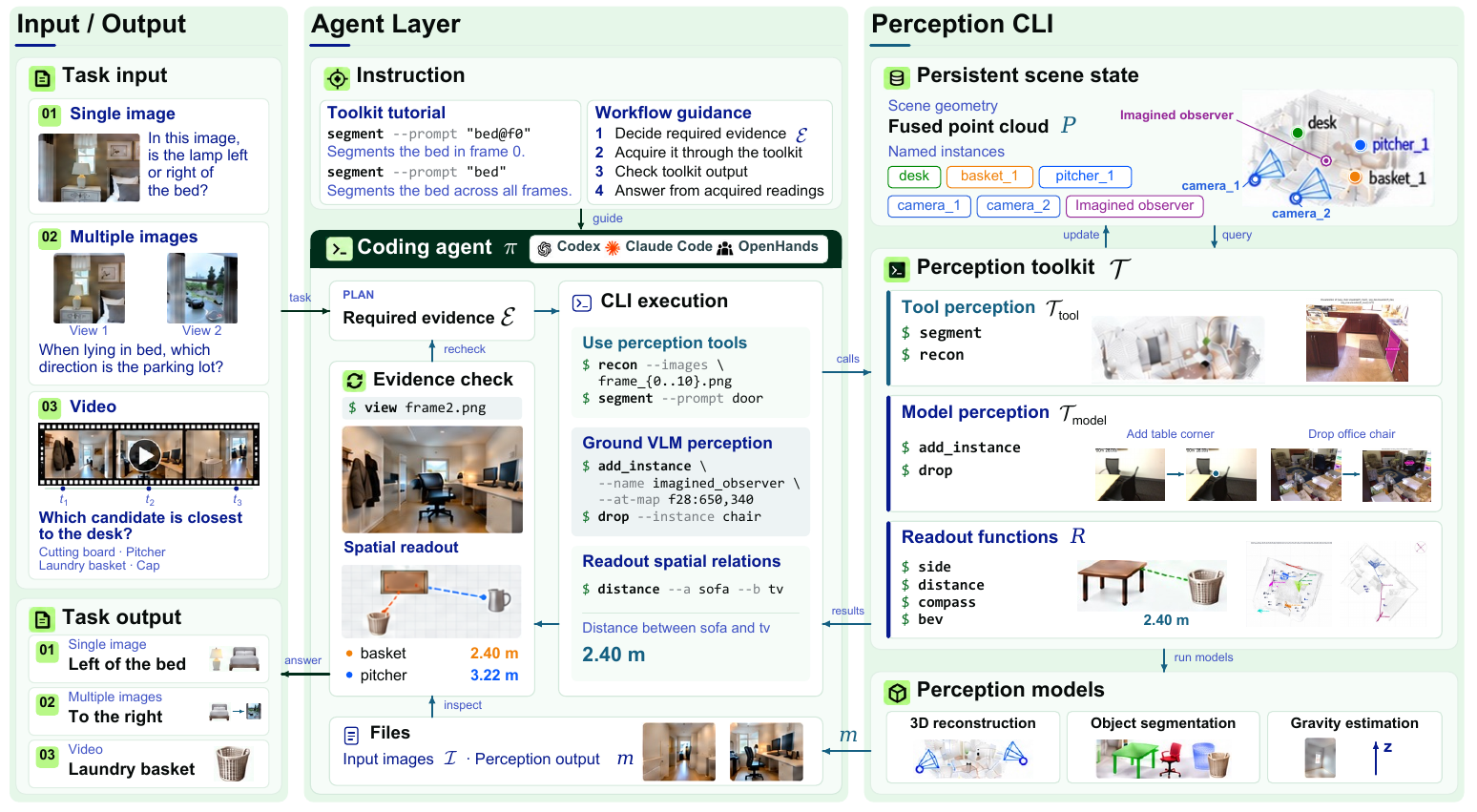}
\end{center}
\caption{Overview of the framework. The agent keeps a coding agent's
full generic freedom over the workspace (viewing images, reading and writing
files, writing and running code) and composes perception through the same
shell; the toolkit registers everything it perceives into the scene memory
and returns interpreted, answer-ready readings.}
\label{fig:system}
\end{figure}

The diagnosis of \S\ref{sec:intro} turns into three requirements that a
system must meet to close the evidence loop, beyond merely possessing tools.
\emph{Acquisition joined with the model's own perception}: what the model
recognizes and what the tools measure must enter one scene.
\emph{Persistent memory}: evidence gathered at one step must remain
intact and addressable many steps later, without the model carrying it
through its own context. \emph{Interpretation on arrival}: a measurement
should arrive in the spatial language the question is asked in.
Existing systems supply acquisition generously, offering a
Turing-complete kernel over perception and geometry libraries, but they
keep the model's own perception in a separate channel and leave memory
and interpretation to the model. \sysname{} inverts that allocation: we
use a general coding agent as the spatial agent, and the design effort
moves into a tool interface that joins the two kinds of perception and
carries memory and interpretation itself. 
Figure~\ref{fig:system} sketches the overall framework.

\subsection{A general coding agent as the spatial agent}
\label{sec:method-overview}

\sysname{} has three parts: a general-purpose coding agent $\pi$ as
planner and executor; a command-line interface (CLI) perception toolkit
$\mathcal{T}$; and a persistent per-episode \emph{scene memory} $S$
that the toolkit reads and writes.
Formally, a spatial question consists of a question $q$ and frames
$\mathcal{I}=\{I_1,\dots,I_n\}$. The agent receives a fixed instruction
$u$, the same for every episode, which holds a tutorial for the
toolkit's commands and a recommended workflow: decide the evidence
$\mathcal{E}$ the question requires, acquire $\mathcal{E}$ through the
toolkit, check the perception tool outputs, and answer from the
acquired readings. At reasoning step $t$ the agent's
context holds the instruction $u$, the question $q$, and the history of
its own chain of thought $z_{<t}$, actions $a_{<t}$, and tool responses
$o_{<t}$; from this context the policy $\pi$ produces the next thought
and action,
\begin{equation}
(z_t,\, a_t) \sim \pi\!\left(\cdot \mid u,\, q,\, z_{<t},\, a_{<t},\, o_{<t}\right),
\qquad
a_t \in \mathcal{A}_{\mathrm{gen}} \cup \mathcal{A}_{\mathrm{perc}},
\label{eq:agent}
\end{equation}
until it commits an answer. The action space includes two classes of
actions, both operated through the CLI. $\mathcal{A}_{\mathrm{gen}}$
holds the general coding agent's original actions, such as running
Linux commands, writing and executing code, and reading files and
images. $\mathcal{A}_{\mathrm{perc}}$ holds the perception toolkit's
actions, such as reconstruction, segmentation, and readouts of spatial
relations, which the following subsections detail. A perception action
returns a full output $m$ and a tool response $o^{\mathrm{perc}}$. The response
$o^{\mathrm{perc}}$ joins the context, while the full output $m$,
together with the original frames $\mathcal{I}$, is kept on disk. The
agent reads these files through the file-reading actions of
$\mathcal{A}_{\mathrm{gen}}$, deciding when and which to read, and what
it reads joins the context as $o^{\mathrm{gen}}$. Across these steps
the model's own work is the two maps $q \mapsto \mathcal{E}$ and
$o^{\mathrm{perc}} \mapsto \mathcal{E}$: deciding what evidence the
question requires, and reading each reply into a piece of that
evidence. Since this design
assumes only the CLI machinery every coding agent provides, our
framework works across general coding agent harnesses
(\S\ref{sec:exp-harness}).

\subsection{Scene memory kept by the environment}
\label{sec:method-memory}

Closing the loop on a multi-view or video question takes many steps, and
the evidence gathered early must still be at hand when the answer is
committed, so some component has to hold it. The model's context window can
serve as that holder, at the price of transcribing every intermediate
result into text and carrying it across a growing transcript.

We let the environment do the remembering instead: the toolkit maintains a
persistent \emph{scene memory}
\begin{equation}
S \;=\; \big(\mathcal{I},\ P,\ \{(\nu_i,\, g_i)\}_{i}\big),
\qquad
g_i \;=\; \big(\mathbf{p}_i,\, E_i,\, T_i\big),
\label{eq:scene}
\end{equation}
holding the raw frames $\mathcal{I}$, the fused metric point cloud $P$ in
a gravity-aligned world frame, and a registry of named instances. Each
instance pairs a name $\nu_i$, usable in later calls, with its geometry
$g_i$: an $xyz$ position $\mathbf{p}_i \in \mathbb{R}^3$, an extent $E_i$
(the subset of $P$ belonging to the instance), and a 6-DoF pose
$T_i \in \mathrm{SE}(3)$; $E_i$ and $T_i$ may each be left empty. Both
the perception toolkit and the model can write to this memory;
\S\ref{sec:method-division} details both routes.

Because the scene memory records each instance's position, extent, and
pose, measuring objects and their relations becomes a lookup, computed
in closed form from the stored geometry. Readout actions are a family
of functions $R_\tau(\nu_1, \dots, \nu_k;\, S_t)$ with
$\tau \in \mathcal{T}$, over one or more
named instances, covering properties of a single instance, such as its
size and elevation, and relations between instances, such as pairwise
distance, relative direction, and motion between frames. A reading is derived fresh
from the stored geometry at each call, so the model can complete a
readout by remembering only the instance names.
Figure~\ref{fig:example} lists what one episode's memory holds.

\subsection{Dividing perception by competence}
\label{sec:method-division}

The evidence loop involves two kinds of perceptual work with opposite
competence profiles. Deciding \emph{what} to measure is semantic work
that VLMs do well, such as identifying which objects the question is
about and phrasing a segmentation prompt. Producing precise masks and
measurements is work that dedicated perception models do better. A
principled interface should cut exactly along this line. Formally, a
perception action pairs a toolkit command $\tau$ with the arguments $x$
the model supplies: the model contributes the map $q \mapsto x$, the
grounding step that carries the evidence $\mathcal{E}$ it conceives into
terms the toolkit operates on, and the toolkit supplies
$(x, S) \mapsto o^{\mathrm{perc}}$. Our interface lets the model write
$x$ directly from earlier replies, $o^{\mathrm{perc}} \mapsto x$. The
call has semantics
\begin{equation}
(o^{\mathrm{perc}}_t,\, S_{t+1}) \;=\; \tau\!\left(x_t;\, S_t\right),
\qquad
\tau \in \mathcal{T} = \mathcal{T}_{\mathrm{tool}} \cup \mathcal{T}_{\mathrm{model}},
\label{eq:call}
\end{equation}
where $S_t$ is the scene memory of \S\ref{sec:method-memory}.

\begin{figure}[t]
\begin{center}
\includegraphics[width=\linewidth]{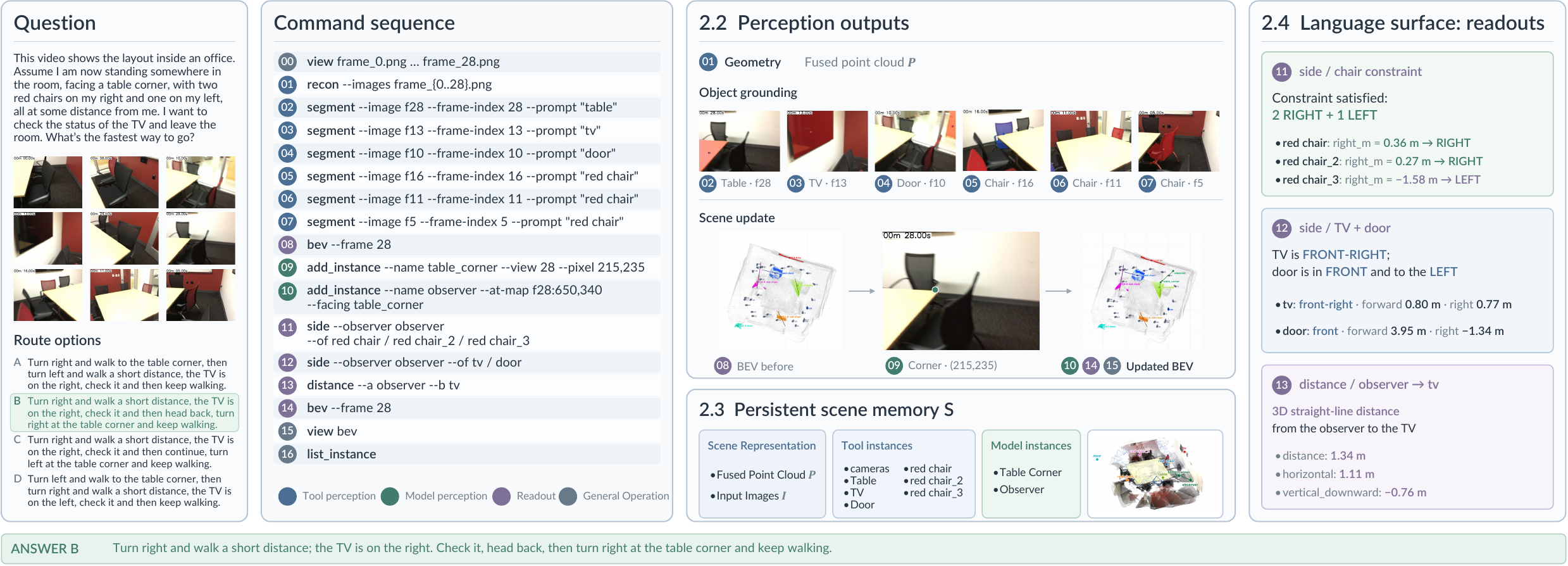}
\end{center}
\caption{An example question used throughout this section. Left to right: the question, the agent's
command sequence colour-coded by who performs the perception, the
outputs those calls produce and the scene memory they write, and the
readouts that answer the question.}
\label{fig:example}
\end{figure}

We build the division into the interface along these two write routes
into the scene memory. \emph{Tool perception}
($\mathcal{T}_{\mathrm{tool}}$) carries the measurement side. The
\code{recon} command fuses all frames into one metric, gravity-aligned
scene and registers every camera in the scene memory; the \code{segment}
command grounds objects to pixels, taking the model's own words as
prompts, so deciding what to look for remains semantic work.
\emph{Model perception} ($\mathcal{T}_{\mathrm{model}}$) enters through
the \code{add-instance} command, which grounds the model's high-level
understanding itself. The model names the new instance, such as an
object it recognizes or an observer the question imagines, and
initializes its position and pose from other named instances or from
pixel locations in the images. The toolkit then treats it like any other
instance, so a construct that exists only in the model's reading of the
question is measured on the same geometric footing as the instances that
tool perception grounds. The same route also runs in reverse: with
\code{drop} the model removes an instance the tools registered wrongly,
such as a mask on a reflection, so its own look at the tool perception outputs corrects
the memory before any reading is taken from it.
Figure~\ref{fig:example} shows model perception and tool perception apart in one episode.
Detailed descriptions of toolkit interfaces are given in Appendix~\ref{sec:appx-toolkit}.

\subsection{The language surface: numbers interpreted on arrival}
\label{sec:method-lastmile}

With acquisition routine and memory kept by the environment, the
remaining break point is how a reading is consumed. Even when the
measurement itself is correct, the model may take it in the wrong
reference frame, invert which object it is measured from, or ignore it
entirely. A displacement vector alone does not say which object it points
from, whether its axes follow gravity or the camera, or which answer
option it favors. That knowledge lives outside the numbers, and the side
that computed the geometry holds it most reliably. In previous
tool-augmented systems~\citep{spatialclaw}, tool calls return bare
values or arrays, leaving the whole step from number to usable spatial
statement to the model. In the failures we observed, the model calls the
right tool and the tool returns the right geometry, yet the final answer
still comes out wrong.

We therefore have every reply interpret its own numbers, handing the
model the values and their plain-language reading together. Each command
distills its full output $m$, against the scene memory $S$, into a
compact structured report
\begin{equation}
o^{\mathrm{perc}} \;=\; c_\tau(m;\, S),
\label{eq:reply}
\end{equation}
with a fixed anatomy: a few typed numeric fields with units, together
with a \emph{conclusion}, one sentence that restates the measurement in
plain spatial language. The conclusion names the entities involved, declares
the frame it speaks in (gravity-aligned, camera, or image plane), and
carries the reading all the way to the ordinary spatial vocabulary an
answer is written in. Figure~\ref{fig:example} gives an example:
the responses describe relation in words first and keeps the numerical
fields that support it, so the model can act on a reading without
re-deriving it. And the complete
measurements and geometry stay in the full output $m$ on disk, where
the model can read them through the same CLI when it needs them.

Together these commitments close the evidence loop. Evidence travels in
rounds,
\begin{center}
\begin{tikzpicture}[font=\normalsize,
  every node/.style={inner sep=2pt},
  fwd/.style={-{Stealth[length=1.8mm]}, semithick},
  back/.style={-{Stealth[length=1.8mm]}, semithick, dashed, black!60},
]
  \node (q) at (0,0) {$q$};
  \node (E) at (1.25,0) {$\mathcal{E}$};
  \node (x) at (2.5,0) {$x$};
  \node (m) at (3.75,0) {$m$};
  \node (o) at (5.15,0) {$o^{\mathrm{perc}}$};
  \node (ans) at (6.9,0) {answer};
  \draw[fwd] (q) -- (E) node[midway, above] {$\scriptstyle\pi$};
  \draw[fwd] (E) -- (x) node[midway, above] {$\scriptstyle\pi$};
  \draw[fwd] (x) -- (m) node[midway, above] {$\scriptstyle S$};
  \draw[fwd] (m) -- (o) node[midway, above] {$\scriptstyle S$};
  \draw[fwd] (o) -- (ans) node[midway, above] {$\scriptstyle\pi$};
  \draw[back] (o) to[bend left=40]
    node[midway, below] {$\scriptstyle\pi$} (x);
  \draw[back] (o.north) to[bend right=45]
    node[midway, above] {$\scriptstyle\pi$} (E.north);
\end{tikzpicture}
\end{center}
where $\pi$ is the VLM, $S$ is the scene memory $q$ is the question, $\mathcal{E}$ the evidence the model decides
the question requires, $x$ the arguments of a perception call, $m$ the
full output the call writes to disk, and $o^{\mathrm{perc}}$ the
interpreted reply; $\pi$ marks the model's steps and $S$ the toolkit's,
computed against the scene memory. The dashed returns close the loop:
the model reads each reply back into its evidence $\mathcal{E}$ and
into the arguments of the next call.

\section{Experiments}
\label{sec:exp}

\subsection{Setup}
\label{sec:exp-setup}

\textbf{Benchmarks.} We evaluate \sysname{} on five benchmarks
covering multi-view and video spatial reasoning. For \emph{multi-view}
reasoning, \textbf{MMSI-Bench}~\citep{yang2026mmsi} provides
multiple images of the same scene and tests whether models integrate
evidence across views for positional relationships, geometric
attributes, motion, and multi-step reasoning;
\textbf{MindCube}~\citep{wang2025mindcube} probes the spatial
mental model built from a few sparse views, asking for relations the
given views never show directly; \textbf{SPAR-Bench}~\citep{zhang2026sparbench}
(used for the harness-swap study) spans a hierarchy of
spatial capabilities from basic metric perception to multi-view
relational reasoning. For \emph{video} reasoning,
\textbf{ReVSI}~\citep{zhang2026revsi} evaluates 3D spatial
reasoning from dynamic observations, emphasizing relations that no
isolated frame reliably shows;
\textbf{MMSI-Video-Bench}~\citep{mmsi-video} extends the MMSI setting
to long videos, with questions on spatial construction, motion
understanding, cross-video memory, and route planning. ReVSI and
SPAR-Bench scores follow the official protocols, macro-averaged over
question types; the others report overall accuracy.

\textbf{Models.} We use Qwen3.8-27B (FP8, served locally) as the primary backbone;
Gemini-3.8-Flash as the stronger backbone for the capability study. The
general coding agent harness is OpenHands~\citep{wang2025openhands} unless stated otherwise, and the
same workspaces and toolkit also run under Codex and Claude Code
(\S\ref{sec:exp-harness}). The perception tools are served by
off-the-shelf models, all frozen: Depth Anything 3~\citep{da3} for the
metric reconstruction, $\pi^3_{\mathrm{mos}}$~\citep{wang2026pi3,zhong2026pi3mos}
for per-pixel motion masks, GeoCalib~\citep{veicht2024geocalib} for the
gravity direction, and SAM 3~\citep{sam3} for promptable
segmentation.

\textbf{Baselines.} We compare \sysname{} with three categories of
baselines: advanced proprietary VLMs, such as GPT-5.4, Grok~4, and the
Gemini series; open-weight general and spatially specialized models,
such as the Qwen series and the SenseNova-SI series; and agentic
systems, S-Agent~\citep{sagent} and SpatialClaw~\citep{spatialclaw}.

\subsection{Main result}
\label{sec:exp-main}

\begin{table}[t]
\caption{MMSI-Bench results by question category (C/O/R denote
camera/object/region).}
\label{tab:mmsi-sota}
\begin{center}
\resizebox{\linewidth}{!}{%
\begin{tabular}{lcccccc|cc|cc|c|c}
\toprule
\multirow{2}{*}{Model} & \multicolumn{6}{c}{Positional Relationship} &
\multicolumn{2}{c}{Geometric Attr.} & \multicolumn{2}{c}{Motion} &
\multirow{2}{*}{MSR} & \multirow{2}{*}{Avg.} \\
\cmidrule(lr){2-7}\cmidrule(lr){8-9}\cmidrule(lr){10-11}
& C-C & O-O & R-R & C-O & O-R & C-R & Meas. & Appr. & Cam. & Obj. & & \\
\midrule
\multicolumn{13}{l}{\emph{Proprietary models}} \\
GPT-5.4        & 41.9 & 33.0 & 35.8 & 49.8 & 42.4 & 68.7 & 54.7 & 37.4 & 28.3 & 40.8 & 36.4 & 41.9 \\
GPT-5.6        & 64.5 & 42.6 & 55.6 & 41.9 & 44.7 & 59.0 & 67.2 & 40.9 & 59.5 & 48.7 & 57.1 & 53.2 \\
GPT-6          & \textbf{73.1} & 61.7 & \textbf{72.8} & 58.1 & 67.1 & 72.3 & \textbf{73.4} & \textbf{69.7} & \textbf{89.2} & 65.8 & \textbf{75.3} & 71.0 \\
Grok 4         & 36.6 & 35.1 & 39.5 & 34.9 & 45.9 & 50.6 & 21.9 & 22.7 & 40.5 & 43.4 & 38.4 & 37.8 \\
Gemini 2.5 Pro & 38.7 & 34.0 & 40.7 & 44.2 & 38.8 & 41.0 & 62.5 & 30.3 & 39.2 & 25.0 & 33.3 & 38.0 \\
Gemini 3 Pro   & 47.3 & 48.9 & 42.0 & 43.0 & 37.6 & 60.2 & 64.1 & 39.4 & 41.9 & 47.4 & 37.9 & 45.2 \\
Gemini 3.8 Flash & 68.8 & 60.6 & 51.9 & 51.2 & 52.9 & 62.7 & 64.1 & 47.0 & 78.4 & 61.8 & 60.1 & 60.0 \\
\midrule
\multicolumn{13}{l}{\emph{Open-weight general models}} \\
Seed 1.6                  & 36.6 & 36.2 & 32.1 & 32.6 & 42.4 & 46.9 & 48.4 & 33.0 & 31.1 & 42.1 & 40.4 & 38.5 \\
InternVL3\_5-8B           & 29.0 & 26.6 & 29.6 & 24.4 & 31.8 & 25.3 & 29.7 & 25.8 & 14.9 & 34.2 & 36.4 & 29.0 \\
SenseNova-U1-8B-MoT       & 46.2 & 41.5 & 29.6 & 58.1 & 38.8 & 63.9 & 43.8 & 21.2 & 25.7 & 31.6 & 26.8 & 38.0 \\
Qwen3-VL-8B-Instruct      & 28.0 & 37.2 & 32.1 & 31.4 & 35.3 & 38.5 & 37.5 & 15.2 & 27.0 & 28.9 & 29.8 & 31.1 \\
Qwen3-VL-8B-Thinking      & 31.2 & 26.6 & 32.1 & 29.1 & 32.9 & 30.1 & 50.0 & 16.7 & 17.6 & 23.7 & 27.3 & 28.6 \\
Qwen3.5-9B                & 34.4 & 36.2 & 34.6 & 39.5 & 38.8 & 54.2 & 56.3 & 28.8 & 36.5 & 26.3 & 28.8 & 36.5 \\
Qwen3-VL-30B-A3B-Thinking & 23.7 & 31.9 & 35.8 & 31.4 & 36.5 & 22.9 & 40.6 & 19.7 & 18.9 & 27.6 & 31.3 & 29.4 \\
Qwen3.8-27B               & 58.1 & 38.3 & 44.4 & 36.0 & 49.4 & 55.4 & 56.2 & 47.0 & 67.6 & 55.3 & 50.5 & 50.4 \\  
\midrule
\multicolumn{13}{l}{\emph{Open-weight spatial models}} \\
SenseNova-SI-1.1-Qwen2.5VL-7B & 51.6 & 29.8 & 32.1 & 50.0 & 29.4 & 42.2 & 37.5 & 28.8 & 23.0 & 34.2 & 18.7 & 32.8 \\
SenseNova-SI-1.1-Qwen3VL-8B   & 44.1 & 38.3 & 33.3 & 65.1 & 38.8 & 59.0 & 48.4 & 24.2 & 29.7 & 34.2 & 22.2 & 38.1 \\
VST-7B-SFT                    & 39.8 & 36.2 & 35.8 & 37.2 & 29.4 & 33.7 & 29.7 & 47.0 & 36.5 & 35.5 & 18.2 & 32.5 \\
\midrule
\multicolumn{13}{l}{\emph{Agentic systems}} \\
S-Agent (Qwen3.8-27B) & 59.1 & 47.9 & 42.0 & 46.5 & 34.1 & 51.8 & 57.8 & 47.0 & 48.6 & 46.1 & 49.5 & 48.3 \\
SpatialClaw (Qwen3.8-27B) & 67.7 & 50.0 & 54.3 & 79.1 & 58.8 & 79.5 & 68.8 & 56.1 & 77.0 & 44.7 & 58.6 & 62.6 \\
SpatialClaw (Gemini-3.8-Flash) & 71.0 & 68.1 & 63.0 & 68.6 & 64.7 & 73.5 & 68.8 & 50.0 & 81.1 & 60.5 & 66.2 & 67.0 \\
\textbf{\sysname{} (Qwen3.8-27B, ours)} & \textbf{73.1}\up{5.4} & 58.5\up{8.5} & 58.0\up{3.7} & 83.7\up{4.6} & 60.0\up{1.2} & 74.7\dn{4.8} & \textbf{73.4}\up{4.6} & 60.6\up{4.5} & 75.7\dn{1.3} & 52.6\up{7.9} & 64.1\up{5.5} & 66.5\up{3.9} \\
\textbf{\sysname{} (Gemini-3.8-Flash, ours)} & 72.0\up{1.0} & \textbf{74.5}\up{6.4} & 67.9\up{4.9} & \textbf{89.5}\up{20.9} & \textbf{68.2}\up{3.5} & \textbf{81.9}\up{8.4} & 71.9\up{3.1} & \textbf{69.7}\up{19.7} & 82.4\up{1.3} & \textbf{68.4}\up{7.9} & 71.2\up{5.0} & \textbf{74.1}\up{7.1} \\
\bottomrule
\end{tabular}%
}
\end{center}
\end{table}

Table~\ref{tab:mmsi-sota} shows the accuracy on MMSI-Bench, overall and
by question type. With the 27B open backbone and no training \sysname{}
is ahead of every agentic baseline and of every proprietary model other
than GPT-6; with Gemini-3.8-Flash it has the highest average in the
table, while GPT-6 keeps the lead on region-to-region relations, camera
motion and multi-step reasoning. The parentheses give the change against
SpatialClaw on the same backbone.

\begin{table}[t]
\caption{ReVSI results by question type. Numerical types (object
counting, absolute distance, object size, room size) use the official
mean relative accuracy; the remaining types are multiple choice.} 
\label{tab:revsi-sota}
\begin{center}
\resizebox{\linewidth}{!}{%
\begin{tabular}{lc|cccc|ccc|c}
\toprule
\multirow{2}{*}{Model} & \multirow{2}{*}{Frames} &
\multicolumn{4}{c}{Numerical Question} &
\multicolumn{3}{c}{Multiple-Choice Question} & \multirow{2}{*}{Avg.} \\
\cmidrule(lr){3-6}\cmidrule(lr){7-9}
& & Obj. Cnt. & Abs. Dist. & Obj. Size & Room Size & Rel. Dist. & Rel. Dir. & Route Plan & \\
\midrule
\multicolumn{10}{l}{\emph{Proprietary models}} \\
GPT-5.2        & 64    & 56.2 & 41.5 & 73.9 & 63.0 & 48.4 & 34.9 & 38.2 & 50.9 \\
GPT-5.4        & 64    & 70.3 & 40.2 & 73.3 & 59.8 & 57.8 & 43.5 & 33.3 & 54.0 \\
GPT-5.6        & 64    & 80.6 & 64.4 & 80.3 & 71.7 & 71.8 & 71.0 & 63.9 & 72.0 \\
Gemini 3 Flash & 1 FPS & 65.7 & 53.1 & 77.6 & 52.8 & 64.6 & 47.9 & 41.8 & 57.6 \\
Gemini 3 Pro   & 1 FPS & 60.1 & 54.7 & 79.3 & 51.9 & 68.1 & 56.0 & 56.4 & 60.9 \\
Gemini 3.8 Flash & 64  & 79.7 & 70.1 & 83.0 & 68.9 & 81.6 & 78.3 & 77.8 & 77.0 \\
GPT-6          & 64    & \textbf{89.2} & 72.5 & 83.4 & 75.0 & 87.4 & 95.7 & 86.1 & 84.2 \\
\midrule
\multicolumn{10}{l}{\emph{Open-weight general models}} \\
Qwen3-VL-8B-Instruct  & 64 & 40.4 & 52.3 & 69.0 & 45.1 & 57.1 & 39.5 & 40.5 & 49.1 \\
Qwen3-VL-32B-Instruct & 64 & 46.9 & 65.0 & 70.4 & 55.8 & 53.8 & 34.0 & 47.3 & 53.3 \\
InternVL3.5-8B        & 64 & 43.3 & 54.6 & 64.2 & 47.6 & 45.0 & 36.3 & 44.4 & 47.9 \\
InternVL3.5-38B       & 64 & 43.8 & 60.6 & 70.2 & 58.4 & 57.4 & 45.9 & 42.7 & 54.1 \\
LLaVA-Video-7B-Qwen2  & 64 & 31.3 & 1.4  & 52.5 & 16.7 & 38.3 & 33.3 & 38.4 & 30.3 \\
LLaVA-Video-72B-Qwen2 & 64 & 40.1 & 29.6 & 59.3 & 27.9 & 39.6 & 24.8 & 43.0 & 37.8 \\
Qwen3.8-27B           & 64 & 63.5 & 52.1 & 71.1 & 53.5 & 64.1 & 63.8 & 61.1 & 61.3 \\
\midrule
\multicolumn{10}{l}{\emph{Spatially specialized models and base models}} \\
Cambrian-S-7B          & 128   & 48.4 & 60.5 & 65.5 & 46.7 & 37.1 & 48.5 & 37.0 & 49.1 \\
Qwen2.5-VL-7B-Instruct & 4 FPS & 36.9 & 15.0 & 49.7 & 29.0 & 31.5 & 29.5 & 36.7 & 32.6 \\
VST-7B-SFT             & 4 FPS & 35.4 & 52.6 & 67.9 & 47.2 & 49.2 & 36.9 & 35.4 & 46.4 \\
Qwen2.5-VL-7B-Instruct & 32    & 34.3 & 21.7 & 45.5 & 35.1 & 32.6 & 33.7 & 34.1 & 33.9 \\
SpaceR-7B (SG-RLVR)    & 32    & 30.7 & 34.5 & 52.0 & 18.6 & 22.8 & 34.5 & 20.2 & 30.5 \\
Qwen2.5-VL-3B-Instruct & 16    & 18.7 & 15.6 & 16.8 & --   & 33.2 & 34.3 & --   & 23.7 \\
Spatial-MLLM-4B-135k   & 16    & 40.7 & 45.3 & 46.8 & --   & 32.3 & 37.4 & --   & 40.5 \\
Spatial-MLLM-4B-820k   & 16    & 41.5 & 40.0 & 53.1 & --   & 30.7 & 39.2 & --   & 40.9 \\
LLaVA-Video-7B-Qwen2   & 32    & 29.9 & 1.5  & 53.0 & 19.3 & 39.1 & 33.8 & 38.8 & 30.8 \\
VLM3R-7B               & 32    & 41.6 & 61.6 & 64.8 & 52.5 & 46.5 & 49.5 & 34.1 & 50.1 \\
\midrule
\multicolumn{10}{l}{\emph{Agentic systems}} \\
S-Agent (Qwen3.8-27B)     & 64 & 63.0 & 58.1 & 60.9 & 57.4 & 63.6 & 69.1 & 44.4 & 59.5 \\
SpatialClaw (Qwen3.8-27B) & 64 & 74.8 & 79.5 & 73.4 & 59.1 & 84.9 & 77.8 & 72.2 & 74.5 \\
SpatialClaw (Gemini-3.8-Flash) & 64 & 80.3 & \textbf{89.9} & \textbf{86.6} & 75.2 & 87.9 & 63.3 & 75.0 & 79.7 \\
\textbf{\sysname{} (Qwen3.8-27B, ours)} & 64 & 75.2\up{0.4} & 76.7\dn{2.8} & 78.7\up{5.3} & 67.8\up{8.7} & 88.3\up{3.4} & 95.2\up{17.4} & 94.4\up{22.2} & 82.3\up{7.8} \\
\textbf{\sysname{} (Gemini-3.8-Flash, ours)} & 64 & 86.8\up{6.5} & 88.2\dn{1.7} & 86.1\dn{0.5} & \textbf{82.2}\up{7.0} & \textbf{95.1}\up{7.2} & \textbf{97.6}\up{34.3} & \textbf{97.2}\up{22.2} & \textbf{90.5}\up{10.8} \\
\bottomrule
\end{tabular}%
}
\end{center}
\end{table}

Table~\ref{tab:revsi-sota} shows the accuracy on ReVSI, overall and by
question type. \sysname{} is ahead of every agentic baseline on both
backbones, and with Gemini-3.8-Flash it has the highest average in the
table. The margins over the specialized baseline are largest on the types
that need cross-object geometry, relative direction and route planning;
the one deficit is absolute distance, which is also where GPT-6 falls
furthest behind.

\begin{table}[t]
\begin{minipage}[t]{0.56\linewidth}
\caption{Comparison with agentic spatial reasoning baselines,
Qwen3.8-27B backbone.}
\label{tab:main}
\begin{center}
\resizebox{\linewidth}{!}{%
\begin{tabular}{lccccc}
\toprule
Method & MMSI & MindCube & SPAR & MMSI-Vid & ReVSI \\
\midrule
pySpatial   & 36.3 & 66.3 & 42.1 & 26.9 & 39.3 \\
SpaceTools  & 43.2 & 58.6 & 55.3 & 39.3 & 64.4 \\
S-Agent     & 48.3 & 75.5 & 63.0 & 37.9 & 59.5 \\
SpatialClaw & 62.6 & 83.2 & 70.2 & 53.1 & 74.5 \\
\sysname{}  & \textbf{66.5}\up{3.9} & \textbf{88.3}\up{5.1} & \textbf{72.4}\up{2.2} & \textbf{56.4}\up{3.3} & \textbf{82.3}\up{7.8} \\
\bottomrule
\end{tabular}%
}
\end{center}
\end{minipage}\hfill
\begin{minipage}[t]{0.41\linewidth}
\caption{\sysname{} under different coding-agent harnesses,
Qwen3.8-27B backbone.}
\label{tab:harness}
\begin{center}
\resizebox{\linewidth}{!}{%
\begin{tabular}{lccc}
\toprule
Benchmark & OpenHands & Codex & Claude Code \\
\midrule
MMSI-Bench & 66.5 & 67.0 & 63.8 \\
SPAR-Bench & 72.4 & 71.8 & 72.8 \\
ReVSI      & 82.3 & 76.6 & 79.1 \\
\bottomrule
\end{tabular}%
}
\end{center}
\end{minipage}
\end{table}

Table~\ref{tab:main} shows the accuracy on all five benchmarks against
the agentic baselines, where \sysname{} is ahead on every one.

\subsection{Ablations}
\label{sec:exp-ablation}

\noindent\textbf{Different general coding agent harness.}
\label{sec:exp-harness}
We rerun the benchmarks with the whole harness replaced by another
production coding agent, Codex or Claude Code, each with its own agent
loop, prompting and context management, driving the identical workspaces
and toolkit. Table~\ref{tab:harness} shows \sysname{} works under every
one. 

\noindent\textbf{Tools against no tools.}
\label{sec:exp-model}
Table~\ref{tab:model} varies the toolkit against the same agent on the
same items. The toolkit is worth a large gain on both benchmarks on
Qwen3.8-27B, and our score with tools keeps rising as the backbone grows
more capable. SpatialClaw scores below ours on every backbone and
benchmark, and its tool gain shrinks far more as the backbone grows. 

\noindent\textbf{Interface components.}
Table~\ref{tab:ablation} removes one commitment of \S\ref{sec:method} at
a time and leaves the rest of the system untouched. Without the scene
memory the agent carries every reading through its own context; without
the interpreted readout the replies come back as bare values; 
without the model perception interface the model cannot add
instances to the scene or drop them from it.
Every removal costs
accuracy, and the ranking matches the diagnosis of \S\ref{sec:intro}: on
the surface only interpretation goes wrong, but underneath the acquired
evidence is never fully used. Removing the interpreted readout costs the
least ($-1.3$ and $-1.2$); removing the scene memory costs more ($-3.5$
and $-2.5$), since readings carried through the context are lost before
the answer is written; removing the model perception interface costs the
most ($-8.8$ and $-5.9$). Closing the loop needs both the interpretation
design and the scene memory.

\begin{table}[t]
\begin{minipage}[t]{0.53\linewidth}
\caption{Tools against no tools on ReVSI and MMSI-Bench, for
two backbones and two harnesses.}
\label{tab:model}
\begin{center}
\resizebox{0.9\linewidth}{!}{%
\begin{tabular}{llcccc}
\toprule
 & & \multicolumn{2}{c}{Qwen3.8-27B} & \multicolumn{2}{c}{Gemini-3.8-Flash} \\
\cmidrule(lr){3-4}\cmidrule(lr){5-6}
Harness & Tools & ReVSI & MMSI & ReVSI & MMSI \\
\midrule
\multirow{2}{*}{SpatialClaw} & no & 61.3 & 50.4 & 77.0 & 60.0 \\
 & with & 74.5 & 62.6 & 79.7 & 67.0 \\
\cmidrule(lr){1-6}
\multirow{2}{*}{general coding agent} & no & 61.8 & 52.6 & 80.9 & 62.9 \\
 & with & \textbf{82.3} & \textbf{66.5} & \textbf{90.5} & \textbf{74.1} \\
\bottomrule
\end{tabular}%
}
\end{center}
\end{minipage}\hfill
\begin{minipage}[t]{0.40\linewidth}
\caption{Ablations on interface design. Qwen3.8-27B backbone.}
\label{tab:ablation}
\begin{center}
\resizebox{0.9\linewidth}{!}{%
\begin{tabular}{lcc}
\toprule
Variant & MMSI & ReVSI \\
\midrule
\sysname{}                    & \textbf{66.5} & \textbf{82.3} \\
\midrule
no scene memory               & 63.0 & 79.8 \\
no interpretation readout     & 65.2 & 81.1 \\
no model perception interface & 57.7 & 76.4 \\
\bottomrule
\end{tabular}%
}
\end{center}
\end{minipage}
\end{table}

\subsection{Qualitative analysis}
\label{sec:exp-qualitative}

\begin{figure}[t]
\begin{center}
\includegraphics[width=\linewidth]{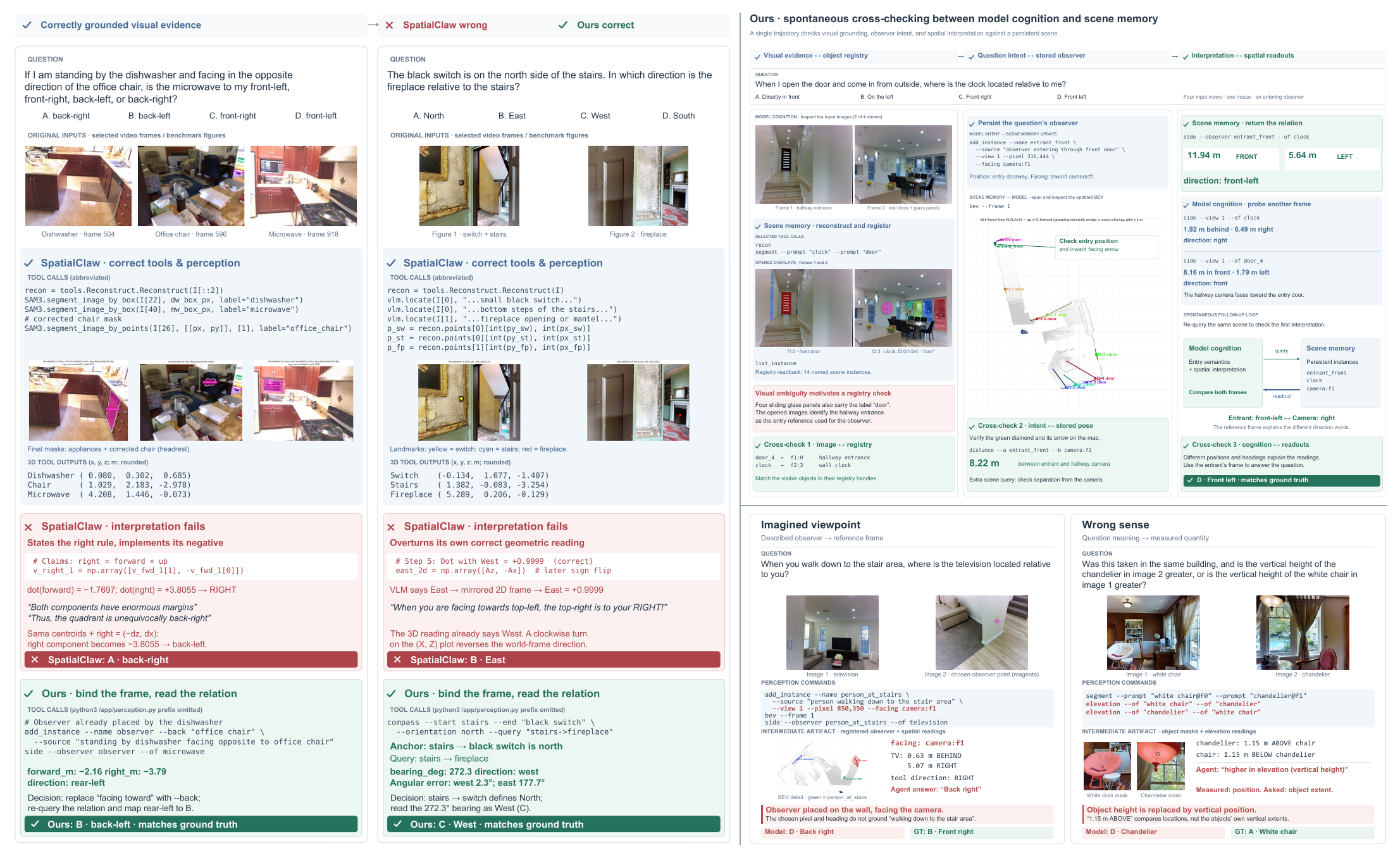}
\end{center}
\caption{Qualitative analysis. Left: two items where SpatialClaw calls the
right tools and gets the right perception, and still answers wrong, with
\sysname{} correct on both (gemini-3.8-flash on both sides).
Top right: \sysname{} cross-checking its own perception against the
scene memory, dropping the wrong masks 
before grounding the observer and reading the answer. Bottom
right: the two remaining grounding failure classes, a wrongly imagined viewpoint
and a quantity read in the wrong sense. Zoom in for better
readability.}
\label{fig:gap}
\end{figure}

\noindent\textbf{Qualitative comparison with SpatialClaw.}
On the same backbone and the same perception backends, the baseline
acquires the evidence and still does not close the loop
(Figure~\ref{fig:gap}, left): the tools return the right masks and the
right positions, and the step from those numbers to the asked relation is
left to the model, which states a right-hand rule correctly and
implements it in reverse, or reaches the correct direction and lets a
second look at the pictures overturn it. Our runs
close the same loop because that step is not left to the model: the
question's premise enters the scene as an oriented instance or a compass
anchor, and the relation comes back in the words the options use.
Seven further items, with the command-line record and the intermediate
images of both agents, are in Appendix~\ref{sec:appx-cases}.

\noindent\textbf{Cross-checking the two kinds of perception.}
The memory closes the loop only if what it holds is right, and the agent
checks that itself. It compares what it sees in a perception output
against what the registry holds and edits the memory when the two disagree
(Figure~\ref{fig:gap}, top right): it drops instances the output shows to
be wrong, and keeps a suspected duplicate once the output shows a genuine
second object. The habit is the coding
agent's inspect-and-revise loop applied to the scene, and it is what joins
the model's perception with the tools' before any reading is taken.

\noindent\textbf{What is left is grounding.}
With acquisition, memory and interpretation carried by the interface, the
errors that remain sit where the question first enters the loop
(Takeaway~3, Figure~\ref{fig:gap}, bottom right). The agent fails to
simulate an imagined viewpoint the question requires, registering an
observer with a borrowed heading, or it measures the right quantity in the
wrong sense, reading an object's elevation above the floor where the
question asks for the object's own height.

\section{Conclusion}
\label{sec:conclusion}

We built a spatial reasoning agent from a stock general coding agent and a
command-line perception toolkit that closes the evidence loop for the
model, joining the tools' measurements with the model's own perception,
keeping every reading in a persistent scene memory, and returning each
reading interpreted in spatial language. Experiments show that with a 27B
open backbone and no training the agent is ahead of every agentic
baseline on all five benchmarks, and that with Gemini-3.8-Flash it has
the highest MMSI-Bench and ReVSI scores of any model we compare against,
including proprietary models such as GPT-6 and GPT-5.6. Its gain from the toolkit remains substantial
on the stronger backbone, where the specialized baseline's shrinks far
more, and the same toolkit works under different coding agent
harnesses.

\subsection*{Reproducibility statement}
Code is available at \url{https://ropedia.github.io/S-Agent-2.0}, together with the
trajectories of all reported runs. Appendix~\ref{sec:appx-toolkit} documents the
toolkit's command-line interface.

\bibliography{references}

\begin{thebibliography}{54}
\providecommand{\natexlab}[1]{#1}

\bibitem[{{Anthropic}(2025)}]{anthropic2025claudecode}
{Anthropic}. 2025.
\newblock Claude 3.7 sonnet and claude code.
\newblock \url{https://www.anthropic.com/news/claude-3-7-sonnet}.
\newblock Accessed: 2026-09-23.

\bibitem[{Brohan et~al.(2023)Brohan, Brown, Carbajal, Chebotar, Chen,
  Choromanski, Ding, Driess, Dubey, Finn, Florence, Fu, Arenas, Gopalakrishnan,
  Han, Hausman, Herzog, Hsu, Ichter, Irpan, Joshi, Julian, Kalashnikov, Kuang,
  Leal, Lee, Lee, Levine, Lu, Michalewski, Mordatch, Pertsch, Rao, Reymann,
  Ryoo, Salazar, Sanketi, Sermanet, Singh, Singh, Soricut, Tran, Vanhoucke,
  Vuong, Wahid, Welker, Wohlhart, Wu, Xia, Xiao, Xu, Xu, Yu, and
  Zitkovich}]{brohan2023rt2}
Anthony Brohan, Noah Brown, Justice Carbajal, Yevgen Chebotar, Xi~Chen,
  Krzysztof Choromanski, Tianli Ding, Danny Driess, Avinava Dubey, Chelsea
  Finn, Pete Florence, Chuyuan Fu, Montse~Gonzalez Arenas, Keerthana
  Gopalakrishnan, Kehang Han, Karol Hausman, Alexander Herzog, Jasmine Hsu,
  Brian Ichter, and 35 others. 2023.
\newblock \href {https://arxiv.org/abs/2307.15818} {Rt-2:
  Vision-language-action models transfer web knowledge to robotic control}.
\newblock \emph{Preprint}, arXiv:2307.15818.

\bibitem[{Caesar et~al.(2020)Caesar, Bankiti, Lang, Vora, Liong, Xu, Krishnan,
  Pan, Baldan, and Beijbom}]{caesar2020nuscenes}
Holger Caesar, Varun Bankiti, Alex~H Lang, Sourabh Vora, Venice~Erin Liong,
  Qiang Xu, Anush Krishnan, Yu~Pan, Giancarlo Baldan, and Oscar Beijbom. 2020.
\newblock nuscenes: A multimodal dataset for autonomous driving.
\newblock In \emph{2020 IEEE/CVF conference on computer vision and pattern
  recognition (CVPR)}, pages 11618--11628. IEEE.

\bibitem[{Cai et~al.(2026)Cai, Wang, Gu, Pu, Xu, Wang, Yin, Yang, Wei, Sun,
  Zhou, Li, Pang, Qian, Wei, Lin, Shi, Deng, Han, Chen, Fan, Deng, Lu, Pan, Li,
  Liu, Wang, Lin, and Yang}]{sensenova_si}
Zhongang Cai, Ruisi Wang, Chenyang Gu, Fanyi Pu, Junxiang Xu, Yubo Wang, Wanqi
  Yin, Zhitao Yang, Chen Wei, Qingping Sun, Tongxi Zhou, Jiaqi Li, Hui~En Pang,
  Oscar Qian, Yukun Wei, Zhiqian Lin, Xuanke Shi, Kewang Deng, Xiaoyang Han,
  and 10 others. 2026.
\newblock Scaling spatial intelligence with multimodal foundation models.
\newblock In \emph{Proceedings of the IEEE/CVF Conference on Computer Vision
  and Pattern Recognition (CVPR)}.

\bibitem[{Carion et~al.(2026)Carion, Gustafson, Hu, Debnath, Hu, Suris
  Coll-Vinent, Ryali, Alwala, Khedr, Huang, Lei, Ma, Guo, Kalla, Marks, Greer,
  Wang, Sun, R\"{a}dle, Afouras, Mavroudi, Xu, Wu, Zhou, Momeni, HAZRA, Ding,
  Vaze, Porcher, Li, Li, Kamath, Cheng, Dollar, Ravi, Saenko, Zhang, and
  Feichtenhofer}]{sam3}
Nicolas Carion, Laura Gustafson, Yuan-Ting Hu, Shoubhik Debnath, Ronghang Hu,
  Didac Suris Coll-Vinent, Chaitanya Ryali, Kalyan~Vasudev Alwala, Haitham
  Khedr, Andrew Huang, Jie Lei, Tengyu Ma, Baishan Guo, Arpit Kalla, Markus
  Marks, Joseph Greer, Meng Wang, Peize Sun, Roman R\"{a}dle, and 19 others.
  2026.
\newblock \href
  {https://proceedings.iclr.cc/paper_files/paper/2026/file/e0982cbc81401df3430ee1ff780dc7a2-Paper-Conference.pdf}
  {Sam 3: Segment anything with concepts}.
\newblock In \emph{International Conference on Learning Representations},
  volume 2026, pages 138846--138923.

\bibitem[{Chen et~al.(2024)Chen, Xu, Kirmani, Ichter, Sadigh, Guibas, and
  Xia}]{chen2024spatialvlm}
Boyuan Chen, Zhuo Xu, Sean Kirmani, Brain Ichter, Dorsa Sadigh, Leonidas
  Guibas, and Fei Xia. 2024.
\newblock Spatialvlm: Endowing vision-language models with spatial reasoning
  capabilities.
\newblock In \emph{Proceedings of the IEEE/CVF conference on computer vision
  and pattern recognition}, pages 14455--14465.

\bibitem[{Chen et~al.(2026{\natexlab{a}})Chen, Uy, Song, Ladhak, Murali, Qu,
  Birchfield, Blukis, and Tremblay}]{chen2026spacetools}
Siyi Chen, Mikaela~Angelina Uy, Chan~Hee Song, Faisal Ladhak, Adithyavairavan
  Murali, Qing Qu, Stan Birchfield, Valts Blukis, and Jonathan Tremblay.
  2026{\natexlab{a}}.
\newblock Spacetools: Tool-augmented spatial reasoning via double interactive
  rl.
\newblock In \emph{Proceedings of the IEEE/CVF Conference on Computer Vision
  and Pattern Recognition}, pages 37109--37120.

\bibitem[{Chen et~al.(2026{\natexlab{b}})Chen, Lu, Zheng, Li, He, Zhou, Shao,
  Zhuang, and Sheng}]{chen2025geometrically}
Zeren Chen, Xiaoya Lu, Zhijie Zheng, Pengrui Li, Lehan He, Yijin Zhou, Jing
  Shao, Bohan Zhuang, and Lu~Sheng. 2026{\natexlab{b}}.
\newblock Geometrically-constrained agent for spatial reasoning.
\newblock In \emph{Proceedings of the IEEE/CVF Conference on Computer Vision
  and Pattern Recognition}, pages 38689--38699.

\bibitem[{Cheng et~al.(2024)Cheng, Yin, Fu, Guo, Yang, Kautz, Wang, and
  Liu}]{cheng2024spatialrgpt}
An-Chieh Cheng, Hongxu Yin, Yang Fu, Qiushan Guo, Ruihan Yang, Jan Kautz,
  Xiaolong Wang, and Sifei Liu. 2024.
\newblock Spatialrgpt: Grounded spatial reasoning in vision-language models.
\newblock \emph{Advances in Neural Information Processing Systems},
  37:135062--135093.

\bibitem[{Cho et~al.(2026)Cho, Hachiuma, Badki, Su, Lee, Song, Liu,
  Radhakrishnan, Kim, Wang, and Chen}]{spatialclaw}
Seokju Cho, Ryo Hachiuma, Abhishek Badki, Hang Su, Byung{-}Kwan Lee, Chan~Hee
  Song, Sifei Liu, Subhashree Radhakrishnan, Seungryong Kim, Yu{-}Chiang~Frank
  Wang, and Min{-}Hung Chen. 2026.
\newblock \href {https://doi.org/10.48550/ARXIV.2606.13673} {Spatialclaw:
  Rethinking action interface for agentic spatial reasoning}.
\newblock \emph{CoRR}, abs/2606.13673.

\bibitem[{Dai et~al.(2026)Dai, Li, Tian, Yao, Dong, Hong, Chen, Liu, Tian,
  Zhang, Wang, Yap, and Liu}]{sagent}
Yalun Dai, Hao Li, Shulin Tian, Runmao Yao, Yuhao Dong, Fangzhou Hong, Zhaoxi
  Chen, Fangfu Liu, Baoliang Tian, Dingwen Zhang, Tao Wang, Kim{-}Hui Yap, and
  Ziwei Liu. 2026.
\newblock \href {https://doi.org/10.48550/ARXIV.2606.20515} {S-agent: Spatial
  tool-use elicits reasoning for spatial intelligence}.
\newblock \emph{CoRR}, abs/2606.20515.

\bibitem[{Driess et~al.(2023)Driess, Xia, Sajjadi, Lynch, Chowdhery, Ichter,
  Wahid, Tompson, Vuong, Yu, Huang, Chebotar, Sermanet, Duckworth, Levine,
  Vanhoucke, Hausman, Toussaint, Greff, Zeng, Mordatch, and
  Florence}]{driess2023palme}
Danny Driess, Fei Xia, Mehdi S.~M. Sajjadi, Corey Lynch, Aakanksha Chowdhery,
  Brian Ichter, Ayzaan Wahid, Jonathan Tompson, Quan Vuong, Tianhe Yu, Wenlong
  Huang, Yevgen Chebotar, Pierre Sermanet, Daniel Duckworth, Sergey Levine,
  Vincent Vanhoucke, Karol Hausman, Marc Toussaint, Klaus Greff, and 3 others.
  2023.
\newblock \href {https://arxiv.org/abs/2303.03378} {Palm-e: An embodied
  multimodal language model}.
\newblock \emph{Preprint}, arXiv:2303.03378.

\bibitem[{Du et~al.(2024)Du, Wu, Li, Huang, and Wei}]{embspatial}
Mengfei Du, Binhao Wu, Zejun Li, Xuan-Jing Huang, and Zhongyu Wei. 2024.
\newblock Embspatial-bench: Benchmarking spatial understanding for embodied
  tasks with large vision-language models.
\newblock In \emph{Proceedings of the 62nd Annual Meeting of the Association
  for Computational Linguistics (Volume 2: Short Papers)}, pages 346--355.

\bibitem[{Fu et~al.(2024)Fu, Hu, Li, Feng, Wang, Lin, Roth, Smith, Ma, and
  Krishna}]{blink}
Xingyu Fu, Yushi Hu, Bangzheng Li, Yu~Feng, Haoyu Wang, Xudong Lin, Dan Roth,
  Noah~A Smith, Wei-Chiu Ma, and Ranjay Krishna. 2024.
\newblock Blink: Multimodal large language models can see but not perceive.
\newblock In \emph{European Conference on Computer Vision}, pages 148--166.
  Springer.

\bibitem[{Geiger et~al.(2012)Geiger, Lenz, and Urtasun}]{geiger2012kitti}
Andreas Geiger, Philip Lenz, and Raquel Urtasun. 2012.
\newblock Are we ready for autonomous driving? the kitti vision benchmark
  suite.
\newblock In \emph{2012 IEEE conference on computer vision and pattern
  recognition}, pages 3354--3361. IEEE.

\bibitem[{Gou et~al.(2024)Gou, Shao, Gong, shen, Yang, Duan, and Chen}]{critic}
Zhibin Gou, Zhihong Shao, Yeyun Gong, yelong shen, Yujiu Yang, Nan Duan, and
  Weizhu Chen. 2024.
\newblock \href
  {https://proceedings.iclr.cc/paper_files/paper/2024/file/fef126561bbf9d4467dbb8d27334b8fe-Paper-Conference.pdf}
  {Critic: Large language models can self-correct with tool-interactive
  critiquing}.
\newblock In \emph{International Conference on Learning Representations},
  volume 2024, pages 57734--57811.

\bibitem[{Gupta and Kembhavi(2023)}]{Gupta2022VisProg}
Tanmay Gupta and Aniruddha Kembhavi. 2023.
\newblock Visual programming: Compositional visual reasoning without training.
\newblock In \emph{2023 IEEE/CVF Conference on Computer Vision and Pattern
  Recognition (CVPR)}, pages 14953--14962. IEEE.

\bibitem[{Jimenez et~al.(2024)Jimenez, Yang, Wettig, Yao, Pei, Press, and
  Narasimhan}]{jimenez2024swebench}
Carlos~E Jimenez, John Yang, Alexander Wettig, Shunyu Yao, Kexin Pei, Ofir
  Press, and Karthik Narasimhan. 2024.
\newblock Swe-bench: Can language models resolve real-world github issues?
\newblock In \emph{International Conference on Learning Representations},
  volume 2024, pages 54107--54157.

\bibitem[{Li et~al.(2026)Li, Li, Wang, Yan, Wu, Zhang, Shen, Lu, Xiao, and
  Zhuang}]{spatialladder}
Hongxing Li, Dingming Li, Zixuan Wang, Yuchen Yan, Hang Wu, Wenqi Zhang,
  Yongliang Shen, Weiming Lu, Jun Xiao, and Yueting Zhuang. 2026.
\newblock Spatialladder: Progressive training for spatial reasoning in
  vision-language models.
\newblock In \emph{International Conference on Learning Representations},
  volume 2026, pages 76566--76592.

\bibitem[{Lin et~al.(2026)Lin, Chen, Liew, Chen, Li, Zhao, Peng, Guo, Zhou,
  Shi, Feng, and Kang}]{da3}
Haotong Lin, Sili Chen, Jun~Hao Liew, Donny~Y. Chen, Zhenyu Li, Yang Zhao, Sida
  Peng, Hengkai Guo, Xiaowei Zhou, Guang Shi, Jiashi Feng, and Bingyi Kang.
  2026.
\newblock \href {https://openreview.net/forum?id=yirunib8l8} {Depth anything 3:
  Recovering the visual space from any views}.
\newblock In \emph{The Fourteenth International Conference on Learning
  Representations}.

\bibitem[{Lin et~al.(2025)Lin, Xu, Zhu, Yang, Cao, Ran, Hu, Zhu, Xie, Long, Hu,
  Lin, Wang, and Pang}]{mmsi-video}
Jingli Lin, Runsen Xu, Shaohao Zhu, Sihan Yang, Peizhou Cao, Yunlong Ran, Miao
  Hu, Chenming Zhu, Yiman Xie, Yilin Long, Wenbo Hu, Dahua Lin, Tai Wang, and
  Jiangmiao Pang. 2025.
\newblock \href {https://doi.org/10.48550/ARXIV.2512.10863} {Mmsi-video-bench:
  {A} holistic benchmark for video-based spatial intelligence}.
\newblock \emph{CoRR}, abs/2512.10863.

\bibitem[{Lu et~al.(2025)Lu, Chen, Liu, Thapa, Boen, and Zou}]{lu2025octotools}
Pan Lu, Bowen Chen, Sheng Liu, Rahul Thapa, Joseph Boen, and James Zou. 2025.
\newblock Octotools: An agentic framework with extensible tools for complex
  reasoning.
\newblock \emph{arXiv preprint arXiv:2502.11271}.

\bibitem[{Luo et~al.(2026)Luo, Zhang, Yong, Dai, Wang, Ran, Shi, Sycara, and
  Xie}]{luo2026pySpatial}
Zhanpeng Luo, Ce~Zhang, Silong Yong, Cunxi Dai, Qianwei Wang, Haoxi Ran, Guanya
  Shi, Katia Sycara, and Yaqi Xie. 2026.
\newblock \href {https://openreview.net/forum?id=yv15C8ql24} {pyspatial:
  Generating 3d visual programs for zero-shot spatial reasoning}.
\newblock In \emph{The Fourteenth International Conference on Learning
  Representations}.

\bibitem[{Marsili et~al.(2025)Marsili, Agrawal, Yue, and
  Gkioxari}]{marsili2025visual}
Damiano Marsili, Rohun Agrawal, Yisong Yue, and Georgia Gkioxari. 2025.
\newblock Visual agentic ai for spatial reasoning with a dynamic api.
\newblock In \emph{2025 IEEE/CVF Conference on Computer Vision and Pattern
  Recognition (CVPR)}, pages 19446--19455. IEEE.

\bibitem[{{OpenAI}(2025)}]{openai2025codex}
{OpenAI}. 2025.
\newblock Introducing codex.
\newblock \url{https://openai.com/index/introducing-codex/}.
\newblock Accessed: 2026-09-23.

\bibitem[{Ouyang et~al.(2025)Ouyang, Liu, Wu, Liu, Zhou, Zhou, Meng, and
  Sun}]{spacer}
Kun Ouyang, Yuanxin Liu, Haoning Wu, Yi~Liu, Hao Zhou, Jie Zhou, Fandong Meng,
  and Xu~Sun. 2025.
\newblock Spacer: Reinforcing mllms in video spatial reasoning.
\newblock \emph{arXiv preprint arXiv:2504.01805}.

\bibitem[{Ropero et~al.(2026)Ropero, Turkoz, Matos, Du, Ruiz, Zhang, Liu, Sun,
  and Wang}]{ropero2026riemind}
Fernando Ropero, Erkin Turkoz, Daniel Matos, Junqing Du, Antonio Ruiz, Yanfeng
  Zhang, Lu~Liu, Mingwei Sun, and Yongliang Wang. 2026.
\newblock Riemind: Geometry-grounded spatial agent for scene understanding.
\newblock \emph{arXiv preprint arXiv:2603.15386}.

\bibitem[{Shen et~al.(2023)Shen, Song, Tan, Li, Lu, and
  Zhuang}]{shen2023hugginggpt}
Yongliang Shen, Kaitao Song, Xu~Tan, Dongsheng Li, Weiming Lu, and Yueting
  Zhuang. 2023.
\newblock Hugginggpt: Solving ai tasks with chatgpt and its friends in hugging
  face.
\newblock \emph{Advances in Neural Information Processing Systems},
  36:38154--38180.

\bibitem[{Shinn et~al.(2023)Shinn, Cassano, Gopinath, Narasimhan, and
  Yao}]{shinn2023reflexion}
Noah Shinn, Federico Cassano, Ashwin Gopinath, Karthik Narasimhan, and Shunyu
  Yao. 2023.
\newblock Reflexion: Language agents with verbal reinforcement learning.
\newblock \emph{Advances in neural information processing systems},
  36:8634--8652.

\bibitem[{Song et~al.(2025)Song, Blukis, Tremblay, Tyree, Su, and
  Birchfield}]{song2025robospatial}
Chan~Hee Song, Valts Blukis, Jonathan Tremblay, Stephen Tyree, Yu~Su, and Stan
  Birchfield. 2025.
\newblock Robospatial: Teaching spatial understanding to 2d and 3d
  vision-language models for robotics.
\newblock In \emph{2025 IEEE/CVF Conference on Computer Vision and Pattern
  Recognition (CVPR)}, pages 15768--15780. IEEE.

\bibitem[{Sur{\'\i}s et~al.(2023)Sur{\'\i}s, Menon, and
  Vondrick}]{surismenon2023vipergpt}
D{\'\i}dac Sur{\'\i}s, Sachit Menon, and Carl Vondrick. 2023.
\newblock Vipergpt: Visual inference via python execution for reasoning.
\newblock In \emph{2023 IEEE/CVF International Conference on Computer Vision
  (ICCV)}, pages 11854--11864. IEEE.

\bibitem[{Veicht et~al.(2024)Veicht, Sarlin, Lindenberger, and
  Pollefeys}]{veicht2024geocalib}
Alexander Veicht, Paul-Edouard Sarlin, Philipp Lindenberger, and Marc
  Pollefeys. 2024.
\newblock Geocalib: Learning single-image calibration with geometric
  optimization.
\newblock In \emph{European Conference on Computer Vision}, pages 1--20.
  Springer.

\bibitem[{Wang et~al.(2025{\natexlab{a}})Wang, Chen, Karaev, Vedaldi,
  Rupprecht, and Novotny}]{vggt}
Jianyuan Wang, Minghao Chen, Nikita Karaev, Andrea Vedaldi, Christian
  Rupprecht, and David Novotny. 2025{\natexlab{a}}.
\newblock Vggt: Visual geometry grounded transformer.
\newblock In \emph{2025 IEEE/CVF Conference on Computer Vision and Pattern
  Recognition (CVPR)}, pages 5294--5306. IEEE.

\bibitem[{Wang et~al.(2025{\natexlab{b}})Wang, Yin, Zhang, Zhang, Wang, Wang,
  Zhang, Chandrasegaran, Liu, Krishna, Xie, Wu, Fei-Fei, and
  Li}]{wang2025mindcube}
Qineng Wang, Baiqiao Yin, Pingyue Zhang, Jianshu Zhang, Kangrui Wang, Zihan
  Wang, Jieyu Zhang, Keshigeyan Chandrasegaran, Han Liu, Ranjay Krishna,
  Saining Xie, jiajun Wu, Li~Fei-Fei, and Manling Li. 2025{\natexlab{b}}.
\newblock Spatial mental modeling from limited views.
\newblock In \emph{The Fourteenth International Conference on Learning
  Representations}.

\bibitem[{Wang et~al.(2024)Wang, Chen, Yuan, Zhang, Li, Peng, and
  Ji}]{wang2024executable}
Xingyao Wang, Yangyi Chen, Lifan Yuan, Yizhe Zhang, Yunzhu Li, Hao Peng, and
  Heng Ji. 2024.
\newblock Executable code actions elicit better llm agents.
\newblock In \emph{International Conference on Machine Learning (ICML)}.

\bibitem[{Wang et~al.(2025{\natexlab{c}})Wang, Li, Song, Xu, Tang, Zhuge, Pan,
  Song, Li, Singh, Tran, Li, Ma, Zheng, Qian, Shao, Muennighoff, Zhang, Hui,
  Lin, Brennan, Peng, Ji, and Neubig}]{wang2025openhands}
Xingyao Wang, Boxuan Li, Yufan Song, Frank~F Xu, Xiangru Tang, Mingchen Zhuge,
  Jiayi Pan, Yueqi Song, Bowen Li, Jaskirat Singh, Hoang Tran, Fuqiang Li, Ren
  Ma, Mingzhang Zheng, Bill Qian, Daniel Shao, Niklas Muennighoff, Yizhe Zhang,
  Binyuan Hui, and 5 others. 2025{\natexlab{c}}.
\newblock \href
  {https://proceedings.iclr.cc/paper_files/paper/2025/file/a4b6ad6b48850c0c331d1259fc66a69c-Paper-Conference.pdf}
  {Openhands: An open platform for ai software developers as generalist
  agents}.
\newblock In \emph{International Conference on Learning Representations},
  volume 2025, pages 65882--65919.

\bibitem[{Wang et~al.(2026)Wang, Zhou, Zhu, Chang, Zhou, Li, Chen, Pang, Shen,
  and He}]{wang2026pi3}
Yifan Wang, Jianjun Zhou, Haoyi Zhu, Wenzheng Chang, Yang Zhou, Zizun Li, Junyi
  Chen, Jiangmiao Pang, Chunhua Shen, and Tong He. 2026.
\newblock \href {https://openreview.net/forum?id=DTQIjngDta}
  {\${\textbackslash}pi{\textasciicircum}3\$: Permutation-equivariant visual
  geometry learning}.
\newblock In \emph{The Fourteenth International Conference on Learning
  Representations}.

\bibitem[{Wu et~al.(2026{\natexlab{a}})Wu, Liu, Hung, and Duan}]{spatial_mllm}
Diankun Wu, Fangfu Liu, Yi-Hsin Hung, and Yueqi Duan. 2026{\natexlab{a}}.
\newblock Spatial-mllm: Boosting mllm capabilities in visual-based spatial
  intelligence.
\newblock \emph{Advances in neural information processing systems},
  38:13569--13597.

\bibitem[{Wu et~al.(2026{\natexlab{b}})Wu, Guan, Feng, Liu, Wu, Wang, Wu, and
  Tan}]{vilasr}
Junfei Wu, Jian Guan, Kaituo Feng, Qiang Liu, Shu Wu, Liang Wang, Wei Wu, and
  Tieniu Tan. 2026{\natexlab{b}}.
\newblock Reinforcing spatial reasoning in vision-language models with
  interwoven thinking and visual drawing.
\newblock \emph{Advances in Neural Information Processing Systems},
  38:143297--143330.

\bibitem[{Yang et~al.(2025)Yang, Yang, Gupta, Han, Fei-Fei, and
  Xie}]{vsi_bench}
Jihan Yang, Shusheng Yang, Anjali~W Gupta, Rilyn Han, Li~Fei-Fei, and Saining
  Xie. 2025.
\newblock Thinking in space: How multimodal large language models see,
  remember, and recall spaces.
\newblock In \emph{2025 IEEE/CVF Conference on Computer Vision and Pattern
  Recognition (CVPR)}, pages 10632--10643. IEEE.

\bibitem[{Yang et~al.(2024)Yang, Jimenez, Wettig, Lieret, Yao, Narasimhan, and
  Press}]{yang2024sweagent}
John Yang, Carlos Jimenez, Alexander Wettig, Kilian Lieret, Shunyu Yao, Karthik
  Narasimhan, and Ofir Press. 2024.
\newblock Swe-agent: Agent-computer interfaces enable automated software
  engineering.
\newblock \emph{Advances in Neural Information Processing Systems},
  37:50528--50652.

\bibitem[{Yang et~al.(2026{\natexlab{a}})Yang, Zhu, Li, Huang, Yan, Zhou, Liu,
  Li, Li, Wang, Lin, and Zhao}]{vst}
Rui Yang, Ziyu Zhu, Yanwei Li, Jingjia Huang, Shen Yan, Siyuan Zhou, Zhe Liu,
  Xiangtai Li, Shuangye Li, Wenqian Wang, Yi~Lin, and Hengshuang Zhao.
  2026{\natexlab{a}}.
\newblock Visual spatial tuning.
\newblock In \emph{European Conference on Computer Vision}, pages 192--211.
  Springer.

\bibitem[{Yang et~al.(2026{\natexlab{b}})Yang, Yang, Huang, II, Yang, Yu, Tong,
  Zheng, Xu, Wang, Fergus, LeCun, Fei-Fei, and Xie}]{cambrian_s}
Shusheng Yang, Jihan Yang, Pinzhi Huang, Ellis L~Brown II, Zihao Yang, Yue Yu,
  Shengbang Tong, Zihan Zheng, Yifan Xu, Muhan Wang, Rob Fergus, Yann LeCun,
  Li~Fei-Fei, and Saining Xie. 2026{\natexlab{b}}.
\newblock \href {https://openreview.net/forum?id=rBFDvZu6pb} {Cambrian-s:
  Towards spatial supersensing in video}.
\newblock In \emph{The Fourteenth International Conference on Learning
  Representations}.

\bibitem[{Yang et~al.(2026{\natexlab{c}})Yang, Xu, Xie, Yang, Li, Lin, Zhu,
  Chen, Duan, Yue, Lin, Wang, and Pang}]{yang2026mmsi}
Sihan Yang, Runsen Xu, Yiman Xie, Sizhe Yang, Mo~Li, Jingli Lin, Chenming Zhu,
  Xiaochen Chen, Haodong Duan, Xiangyu Yue, Dahua Lin, Tai Wang, and Jiangmiao
  Pang. 2026{\natexlab{c}}.
\newblock Mmsi-bench: A benchmark for multi-image spatial intelligence.
\newblock In \emph{International Conference on Learning Representations},
  volume 2026, pages 157051--157088.

\bibitem[{Yang et~al.(2023)Yang, Li, Wang, Lin, Azarnasab, Ahmed, Liu, Liu,
  Zeng, and Wang}]{mm_react}
Zhengyuan Yang, Linjie Li, Jianfeng Wang, Kevin Lin, Ehsan Azarnasab, Faisal
  Ahmed, Zicheng Liu, Ce~Liu, Michael Zeng, and Lijuan Wang. 2023.
\newblock Mm-react: Prompting chatgpt for multimodal reasoning and action.
\newblock \emph{arXiv preprint arXiv:2303.11381}.

\bibitem[{Yao et~al.(2023)Yao, Zhao, Yu, Du, Shafran, Narasimhan, and
  Cao}]{react}
Shunyu Yao, Jeffrey Zhao, Dian Yu, Nan Du, Izhak Shafran, Karthik~R Narasimhan,
  and Yuan Cao. 2023.
\newblock \href {https://openreview.net/forum?id=WE_vluYUL-X} {React:
  Synergizing reasoning and acting in language models}.
\newblock In \emph{The Eleventh International Conference on Learning
  Representations}.

\bibitem[{Ye et~al.(2025)Ye, Wang, Sun, Chandrasegaran, Durante, Eyzaguirre,
  Bisk, Niebles, Adeli, Fei-Fei et~al.}]{tstar}
Jinhui Ye, Zihan Wang, Haosen Sun, Keshigeyan Chandrasegaran, Zane Durante,
  Cristobal Eyzaguirre, Yonatan Bisk, Juan~Carlos Niebles, Ehsan Adeli,
  Li~Fei-Fei, and 1 others. 2025.
\newblock Re-thinking temporal search for long-form video understanding.
\newblock In \emph{2025 IEEE/CVF Conference on Computer Vision and Pattern
  Recognition (CVPR)}, pages 8579--8591. IEEE.

\bibitem[{Zhang et~al.(2023)Zhang, Li, and Bing}]{videollama}
Hang Zhang, Xin Li, and Lidong Bing. 2023.
\newblock Video-llama: An instruction-tuned audio-visual language model for
  video understanding.
\newblock In \emph{Proceedings of the 2023 conference on empirical methods in
  natural language processing: system demonstrations}, pages 543--553.

\bibitem[{Zhang et~al.(2026{\natexlab{a}})Zhang, Chen, Xu, Huang, Mei, Chen,
  Zhou, Yuan, Cai, Huang, Quan, Xu, and Zhang}]{zhang2026sparbench}
Jiahui Zhang, Yurui Chen, Yueming Xu, Ze~Huang, Jilin Mei, Chunhui Chen,
  Yanpeng Zhou, Yu-Jie Yuan, Xinyue Cai, Guowei Huang, Xingyue Quan, Hang Xu,
  and Li~Zhang. 2026{\natexlab{a}}.
\newblock From flatland to space: Teaching vision-language models to perceive
  and reason in 3d.
\newblock \emph{Advances in Neural Information Processing Systems}, 38.

\bibitem[{Zhang et~al.(2025)Zhang, Zhang, Li, Zeng, Yang, Zhang, Wang, Tan, Li,
  and Liu}]{longva}
Peiyuan Zhang, Kaichen Zhang, Bo~Li, Guangtao Zeng, Jingkang Yang, Yuanhan
  Zhang, Ziyue Wang, Haoran Tan, Chunyuan Li, and Ziwei Liu. 2025.
\newblock \href {https://openreview.net/forum?id=30RAWQVGlx} {Long context
  transfer from language to vision}.
\newblock \emph{Transactions on Machine Learning Research}.

\bibitem[{Zhang et~al.(2026{\natexlab{b}})Zhang, Chen, Tan, Mao, Chen, and
  Chang}]{zhang2026revsi}
Yiming Zhang, Jiacheng Chen, Jiaqi Tan, Yongsen Mao, Wenhu Chen, and Angel~X
  Chang. 2026{\natexlab{b}}.
\newblock Revsi: Rebuilding visual spatial intelligence evaluation for accurate
  assessment of vlm 3d reasoning.
\newblock In \emph{Forty-third International Conference on Machine Learning}.

\bibitem[{Zhang et~al.(2026{\natexlab{c}})Zhang, Wu, Jia, Wang, Zhang, Li, Ran,
  Zhang, Sun, Peng, Yin, Wang, and Lu}]{zhang2026think3d}
Zaibin Zhang, Yuhan Wu, Lianjie Jia, Yifan Wang, Zhongbo Zhang, Yijiang Li,
  Binghao Ran, Fuxi Zhang, Zhuohan Sun, Yizhuang Peng, Zhenfei Yin, Lijun Wang,
  and Huchuan Lu. 2026{\natexlab{c}}.
\newblock \href {https://arxiv.org/abs/2601.13029} {Think3d: Thinking with
  space for spatial reasoning}.
\newblock \emph{Preprint}, arXiv:2601.13029.

\bibitem[{Zhao et~al.(2025)Zhao, Zhang, Lin, Li, Wu, Zhang, and
  Wei}]{zhao2025pyvision}
Shitian Zhao, Haoquan Zhang, Shaoheng Lin, Ming Li, Qilong Wu, Kaipeng Zhang,
  and Chen Wei. 2025.
\newblock Pyvision: Agentic vision with dynamic tooling.
\newblock \emph{arXiv preprint arXiv:2507.07998}.

\bibitem[{Zhong et~al.(2026)Zhong, Jin, Popovic, Behley, and
  Stachniss}]{zhong2026pi3mos}
Xingguang Zhong, Liren Jin, Marija Popovic, Jens Behley, and Cyrill Stachniss.
  2026.
\newblock Dynamic visual slam using a general 3d prior.
\newblock In \emph{Proceedings of the IEEE/CVF Conference on Computer Vision
  and Pattern Recognition}, pages 21805--21815.

\end{thebibliography}
\bibliographystyle{ropedia}

\appendix
\clearpage

\section{Related Work}
\label{sec:related}

\paragraph{Spatial reasoning in vision-language models.}
Spatial competence remains one of the weaker abilities of vision-language
models~\citep{chen2024spatialvlm,cheng2024spatialrgpt,song2025robospatial},
and most of the response has been to put the capability into the weights.
One line distills 3D annotations into spatial instruction data at scale, as
in SpatialVLM~\citep{chen2024spatialvlm},
SpatialRGPT~\citep{cheng2024spatialrgpt}, Cambrian-S~\citep{cambrian_s},
and SenseNova-SI~\citep{sensenova_si}. A second line reshapes the backbone
itself with geometry-aware modelling or spatially targeted tuning, as in
Spatial-MLLM~\citep{spatial_mllm} and VST~\citep{vst}. A third line shapes
the training signal through reinforcement learning, verifiable rewards, or
curricula, as in SpaceR~\citep{spacer}, ViLaSR~\citep{vilasr},
MindCube~\citep{wang2025mindcube}, and
SpatialLadder~\citep{spatialladder}. These models answer in a single
forward pass out of what training left in their parameters, taking no
measurement of the scene in front of them, so a new perception backend or a
shifted task distribution asks for another round of training. \sysname{}
leaves the backbone untouched and moves the work to inference time, where
the agent acquires scene evidence on demand and every reading it acts on
stays grounded in the tools' geometry.

\paragraph{Tool-augmented visual agents and code as action.}
Tool-use agents interleave reasoning with calls to external modules,
following the template that ReAct~\citep{react} set on the language side.
On the visual side, VisProg~\citep{Gupta2022VisProg} and
ViperGPT~\citep{surismenon2023vipergpt} synthesize a whole program in one
pass and then execute it, while MM-ReAct~\citep{mm_react},
HuggingGPT~\citep{shen2023hugginggpt}, and
OctoTools~\citep{lu2025octotools} dispatch requests turn by turn through a
structured tool menu, and PyVision~\citep{zhao2025pyvision} writes its
tooling as it goes. CodeAct~\citep{wang2024executable} compared action
formats directly and found executable code to work better than JSON or free
text for general-purpose agents, which is the format a coding agent uses
natively. The tools all of these systems reach for target 2D recognition,
so their controllers are never asked to hold a metric 3D scene across views
and time, which is what a spatial question demands.

\paragraph{Agentic spatial reasoning.}
A recent line brings tool use to spatial questions.
VADAR~\citep{marsili2025visual} synthesizes a task-specific Pythonic API
and then a program over it, and pySpatial~\citep{luo2026pySpatial} composes
reconstruction, camera-pose recovery, and novel-view synthesis into a 3D
visual program. GCA~\citep{chen2025geometrically} splits the model into a
part that formalizes the question as geometric constraints and a part that
executes tool calls within them, RieMind~\citep{ropero2026riemind} queries
an explicit 3D scene graph through typed geometric operations, and
SpaceTools~\citep{chen2026spacetools} fine-tunes a VLM to coordinate a
fixed tool set through supervised demonstrations followed by interactive
reinforcement learning. Think3D~\citep{zhang2026think3d} selects camera
viewpoints, renders a reconstructed point cloud from them, and reasons over
the rendered images. Closest to us, S-Agent~\citep{sagent} schedules 3D
operations across the views and time steps of multi-view sets and long
videos, and SpatialClaw~\citep{spatialclaw} drives the VLM through a
hand-designed plan, act, and reflect loop over a persistent code kernel
stocked with 3D perception and geometry utilities. What these systems share
is a controller written for the purpose and a design effort spent on
acquisition, on ever richer perception and geometry libraries, while
remembering and interpreting the evidence is left to the model and the
model's own perception stays in a separate channel. \sysname{} writes no
controller and spends its design effort on closing the evidence loop at
the interface: the model's own perception enters the same scene as the
tools' measurements, every reading persists in a scene memory that later
calls query by name, and each reading comes back already interpreted in
spatial language.

\paragraph{General-purpose coding agents.}
Software-engineering agents such as SWE-agent~\citep{yang2024sweagent} and
OpenHands~\citep{wang2025openhands}, developed against benchmarks such as
SWE-bench~\citep{jimenez2024swebench}, have matured into harnesses that
supply a shell, a file system, code execution, long-horizon context
management, and a tool-calling loop that the backbone models are trained
to follow. \sysname{} adopts such a harness unchanged and adds perception
as ordinary command-line tools, which is what lets the same method run
under more than one of them (\S\ref{sec:exp-harness}). Two habits of
these agents carry over to spatial reasoning: viewing an image is an
ordinary step in the same stream as the tool replies, and inspecting an
output and revising it is a reflex; together they are what joins the
model's own perception with the tools' before a reading is taken
(\S\ref{sec:exp-qualitative}).

\paragraph{Long-video and multi-view perception.}
Two families of methods supply the visual evidence these questions need.
Frame-compression methods sample, retrieve, or summarize a limited set of
frames for a long-context VLM, as in Video-LLaMA~\citep{videollama},
LongVA~\citep{longva}, and T$^\star$~\citep{tstar}, which keeps the context
tractable at the risk of dropping the frames a question turns on.
Reconstruction-first methods build an explicit 3D representation with
multi-view geometry or feed-forward models such as VGGT~\citep{vggt} and
Depth Anything 3~\citep{da3}, which grounds the geometry well while paying
the reconstruction cost on every question. Both families hand their output
to the model as fixed context, leaving association across views and metric
comparison to implicit reasoning. \sysname{} uses reconstruction and
promptable segmentation backends of this kind as tool perception
(\S\ref{sec:method-division}), with the agent deciding per question which
of them to call and what to ground; what they measure and what the model
registers itself land in one scene memory, and every reading comes back
interpreted.

\paragraph{Spatial reasoning benchmarks.}
Evaluation has moved from single-image probes such as BLINK~\citep{blink}
and EmbSpatial~\citep{embspatial} to settings where the evidence is spread
over many observations. MMSI-Bench~\citep{yang2026mmsi} asks for evidence
integrated across images of one scene, MindCube~\citep{wang2025mindcube}
probes the mental model built from a few sparse views, and
SPAR-Bench~\citep{zhang2026sparbench} spans a hierarchy from metric
perception to multi-view relational reasoning. For video,
VSI-Bench~\citep{vsi_bench} opened the setting,
ReVSI~\citep{zhang2026revsi} rebuilds its evaluation for a more accurate
assessment of 3D reasoning, and MMSI-Video-Bench~\citep{mmsi-video} extends
the MMSI setting to long videos with questions on spatial construction,
motion, cross-video memory, and route planning. We evaluate on five of
these (\S\ref{sec:exp-setup}).

\section{Toolkit interface}
\label{sec:appx-toolkit}

The toolkit is one executable, \code{python3 /app/perception.py
<command> [flags]}, run from the agent's shell like any other program.
Every command reads and writes the scene memory $S$ of
\S\ref{sec:method-memory}; its reply is a JSON document on standard
output, and every image it renders (segmentation overlays, top-down maps)
is written to disk for the agent to open with its ordinary file viewer.
The instruction $u$ is a tutorial over these commands; this appendix lists
the interface itself. Table~\ref{tab:cli} gives every command the agent is
taught, grouped as in \S\ref{sec:method-division}.

\begin{table}[tp]
\caption{The toolkit's command-line interface. Flags marked $^{+}$ are
repeatable. $P$, $\mathcal{I}$ and the instance registry are the
components of the scene memory $S$ (Eq.~\ref{eq:scene}).}
\label{tab:cli}
\begin{center}
\footnotesize
\setlength{\tabcolsep}{3pt}
\begin{tabular}{>{\raggedright\arraybackslash}p{2.4cm}>{\raggedright\arraybackslash}p{3.7cm}>{\raggedright\arraybackslash}p{7.0cm}}
\toprule
Command & Arguments & Effect on $S$ and reply \\
\midrule
\multicolumn{3}{l}{\emph{Tool perception} ($\mathcal{T}_{\mathrm{tool}}$)} \\
\texttt{recon} & none (all frames under \texttt{/app}) & Runs Depth Anything 3~\citep{da3} on all frames for metric depth and camera poses, GeoCalib~\citep{veicht2024geocalib} for the gravity direction and $\pi^3_{\mathrm{mos}}$~\citep{wang2026pi3,zhong2026pi3mos} for per-pixel motion masks; fuses them into the metric, gravity-aligned point cloud $P$ and registers \texttt{camera:f<N>} with its pose and \texttt{scene}. Returns the instance list, with an \texttt{overlap\_warning} naming the seam when two views share almost no covisible surface. \\
\texttt{segment} & \texttt{--prompt TEXT[@f<N>]}$^{+}$, \texttt{--max-masks K} & Runs SAM 3~\citep{sam3} with each text prompt on the named frame or on all frames, lifts every mask onto $P$ and registers each label as an instance with extent $E_i$ (same-label regions within 0.75 m merge), tagged \texttt{static} or \texttt{moving} from the motion masks. Returns, per frame, the overlay PNG path and the region ids with label and centroid, then the updated registry. \\
\midrule
\multicolumn{3}{l}{\emph{Model perception} ($\mathcal{T}_{\mathrm{model}}$)} \\
\texttt{add\_instance} & \texttt{--name}, \texttt{--source}; position \texttt{--at-map f<N>:PX,PY} $|$ \texttt{--view N --pixel X,Y} $|$ \texttt{--at-camera N}; orientation \texttt{--facing V} $|$ \texttt{--primary A=V --secondary A=V} & Registers an instance at the chosen $\mathbf{p}_i$, with pose $T_i$ when an orientation is given; called again with the same name and only orientation flags, re-orients it in place. Returns the updated registry. \\
\texttt{drop} & \texttt{--instance f<N>:<id>}$^{+}$ & Removes those instances permanently. Returns the registry after the drop. \\
\midrule
\multicolumn{3}{l}{\emph{Readouts} $R_\tau$ (closed form over the stored geometry)} \\
\texttt{list\_instance} & none & No change. Returns each instance's name, source, and whether it is oriented. \\
\texttt{bev} & \texttt{--frame N}, \texttt{--erase f<M>:X}$^{+}$ & No change. Returns a top-down map PNG in camera $N$'s ground frame: static points of all frames plus what frame $N$ saw, cameras as wedges, instances as labelled dots, added instances as diamonds with their facing arrow, and the compass rose once anchored. \\
\texttt{distance} & \texttt{--a X --b Y}, \texttt{--known D} & \texttt{--known} stores the stated distance as the scale calibration. Returns straight-line, horizontal and vertical metres and a conclusion. \\
\texttt{depth} & \texttt{--view N} $|$ \texttt{--observer X}; \texttt{--of Y}$^{+}$; \texttt{--known D} & \texttt{--known} calibrates scale. Returns each target's signed depth along the observer's forward axis, a nearest-to-farthest \texttt{order}, and a conclusion. \\
\texttt{side} & \texttt{--view N} $|$ \texttt{--observer X}; \texttt{--of Y}$^{+}$ & No change. Returns, per target, \texttt{forward\_m}, \texttt{right\_m}, \texttt{up\_m} in the observer's frame (a camera's is its gravity-aligned ground heading), the three axis verdicts, the eight-way \texttt{direction}, and a conclusion. \\
\texttt{elevation} & \texttt{--of REF --of Y}$^{+}$ & No change. Returns each target's signed height above the reference along gravity, a lowest-to-highest \texttt{order}, and a conclusion. \\
\texttt{motion} & \texttt{--of A --of B} & No change. For two cameras, returns the translation in the first camera's ground frame, the turn and tilt, and the path of the raw extrinsics; for one object in two frames, its shift as seen from the first camera, its \texttt{rotation} about the vertical and \texttt{image\_rotation} in the image plane; each with a conclusion. \\
\texttt{compass} & \texttt{--start A --end B --orientation DIR}, \texttt{--query "A->B"}$^{+}$, \texttt{--start-plane}, \texttt{--end-plane} & The anchor persists the compass mapping for later calls and maps. Returns the \texttt{direction} of each query, the eight compass words ranked by fit (\texttt{nearest}), and a conclusion that, for an object-to-camera-plane query, states which side of the object that view sees. \\
\bottomrule
\end{tabular}
\end{center}
\end{table}

\textbf{Instance names.} Every argument that refers to something in the
scene is a name. \code{recon} registers \code{camera:f<N>} for the camera
of frame $N$ (position and orientation) and \code{scene} for the room
centre. \code{segment} registers each prompt's label as an instance
(\code{bed}, \code{desk}); a second, separate object under the same label
becomes \code{<label>\_2}, and \code{f<N>:<label-or-id>} names one frame's
view of an object, by label or by the id stamped on that frame's overlay.
\code{add\_instance} names are chosen by the model. Two forms exist for
\code{compass} only: \code{camplane:f<N>}, the image plane of camera $N$,
and \code{<name>|plane}, the plane fitted through an instance's points.

\textbf{Positions and orientations for \code{add\_instance}.} A position
comes from one of three sources: \code{--at-map f<N>:PX,PY}, a pixel of
the top-down map rendered from frame $N$ (height filled in from the
reconstruction at that spot); \code{--view N --pixel X,Y}, a pixel of the
original frame lifted to its 3D point; or \code{--at-camera N}, the
position of camera $N$. An orientation is a primary axis taken exactly and
a secondary direction adjusted to sit orthogonal to it; the axes are the
instance's own (forward/back, right/left, top/bottom) and a direction
value is any existing instance, the literal \code{gravity}, or
\code{bev:f<N>:START:END}, an arrow between two points on the map from
frame $N$. \code{--facing V} is the upright shorthand (gravity fills the
vertical, \code{V} gives the heading), \code{--primary AXIS=V --secondary
AXIS=V} the general form. A second \code{add\_instance} call with the same
\code{--name} and only orientation flags orients the existing instance in
place.

\textbf{Reply anatomy.} Every reply has the same shape: a status flag,
typed numeric fields in metres on the reconstruction's own scale (a
\code{units} field says whether the scale has been calibrated with
\code{--known}), and a \code{conclusion} sentence per target that restates
the measurement in spatial language.
The \code{side} reply of the episode in Figure~\ref{fig:gap} (bottom
right) reads, verbatim:
{\small
\begin{verbatim}
$ python3 /app/perception.py side --observer sitting_person --of desk
{
  "ok": true,
  "observer": "sitting_person",
  "sides": [
    {
      "target": "desk",
      "forward_m": 1.04, "right_m": -1.03, "up_m": 0.21,
      "front_or_behind": "front", "left_or_right": "left",
      "above_or_below": "above", "direction": "front-left",
      "conclusion": "desk is 1.04 m in FRONT of sitting_person, 1.03 m
        to its LEFT and 0.21 m ABOVE it -- overall to its FRONT-LEFT"
    }
  ],
  "units": "metres on the reconstruction's own metric scale
    (uncalibrated)"
}
\end{verbatim}
}
The executable also exposes lower-level commands (\code{detect},
\code{measure}, \code{match}, \code{egomotion}, \code{gravity},
\code{render}) that the instruction tells the agent not to use.

\section{Toolkit usage}
\label{sec:appx-usage}

Table~\ref{tab:usage} counts, over every episode of three full-benchmark
runs of \sysname{} under OpenHands, how often the agent issues each
command of Table~\ref{tab:cli} and how often it opens each kind of image.
Building the scene is universal: \code{recon}, \code{segment} and
\code{bev} appear in over ninety percent of episodes on every run. The
readouts follow the questions: MMSI-Bench, whose items ask about
relations between views and viewpoints, draws \code{side}, \code{motion}
and \code{compass}; ReVSI, whose items ask about distances, sizes and
counts in a video, draws \code{distance} and \code{list\_instance}
instead. The model perception interface is used in a quarter to a half
of the episodes, and \code{drop} in a fifth to two fifths, so registry
edits are a routine part of an episode rather than an exception. The
lower-level commands the instruction reserves are called in under two
percent of all calls.

\begin{table}[tp]
\caption{Toolkit usage per episode. Calls per episode and, in
parentheses, the share of episodes that issue the command at least once.
Image views count the agent's own \code{view} calls on benchmark frames,
segmentation overlays and top-down maps. MMSI-Bench and ReVSI, 1000 items
each.}
\label{tab:usage}
\begin{center}
\footnotesize
\begin{tabular}{lccc}
\toprule
 & Qwen3.8-27B, MMSI & Gemini-3.8-Flash, MMSI & Gemini-3.8-Flash, ReVSI \\
\midrule
\texttt{recon}          & 1.00 (100\%) & 1.00 (100\%) & 1.00 (100\%) \\
\texttt{segment}        & 1.75 (93\%) & 1.47 (92\%) & 2.25 (100\%) \\
\texttt{add\_instance}  & 1.47 (36\%) & 0.72 (26\%) & 1.48 (48\%) \\
\texttt{drop}           & 0.37 (26\%) & 0.27 (21\%) & 0.53 (40\%) \\
\texttt{list\_instance} & 0.29 (24\%) & 0.52 (45\%) & 1.11 (91\%) \\
\texttt{bev}            & 1.66 (94\%) & 1.67 (94\%) & 1.74 (98\%) \\
\texttt{distance}       & 1.53 (46\%) & 0.37 (17\%) & 2.86 (75\%) \\
\texttt{depth}          & 0.18 (12\%) & 0.02 (2\%) & 0.01 (1\%) \\
\texttt{side}           & 1.66 (62\%) & 1.06 (53\%) & 0.50 (29\%) \\
\texttt{elevation}      & 0.17 (7\%) & 0.11 (7\%) & 0.11 (8\%) \\
\texttt{motion}         & 1.08 (73\%) & 0.91 (66\%) & 0.06 (3\%) \\
\texttt{compass}        & 0.55 (25\%) & 0.69 (25\%) & 0.01 (0\%) \\
\midrule
frame views             & 2.23 & 2.52 & 11.43 \\
overlay views           & 2.12 & 1.26 & 2.70 \\
map views               & 1.59 & 1.40 & 1.36 \\
\bottomrule
\end{tabular}
\end{center}
\end{table}

\section{SpatialClaw with our readouts}
\label{sec:appx-sc-readouts}

To separate what the readouts contribute from what the rest of the
interface contributes, we add our readout commands to SpatialClaw's
kernel as plain functions and leave its harness, prompts and perception
backends untouched. The transplanted set is \code{distance},
\code{depth}, \code{side}, \code{elevation}, \code{motion} and
\code{compass}; each returns the same interpreted reply as in
Table~\ref{tab:cli}, packed into a Python dictionary with the same keys
(the numeric fields, the per-target \code{conclusion} and \code{units}),
so the model reads it in its kernel exactly as our agent reads the JSON on
standard output. \code{list\_instance} is excluded because it reads
the scene memory that SpatialClaw does not keep, and \code{bev} because
SpatialClaw already renders a top-down map of its own.
Table~\ref{tab:sc-readouts} compares the three systems on the same items
as Table~\ref{tab:model}, all on the Qwen3.8-27B backbone.

\begin{table}[tp]
\caption{SpatialClaw with our readouts transplanted into its kernel,
against SpatialClaw and \sysname{}. Qwen3.8-27B backbone.}
\label{tab:sc-readouts}
\begin{center}
\begin{tabular}{lcc}
\toprule
System & ReVSI & MMSI \\
\midrule
SpatialClaw                  & 74.5 & 62.6 \\
SpatialClaw + our readouts   & 75.5   & 62.9   \\
\sysname{}                   & \textbf{82.3} & \textbf{66.5} \\
\bottomrule
\end{tabular}
\end{center}
\end{table}

\section{Additional comparisons with SpatialClaw}
\label{sec:appx-cases}

\begin{figure}[p]
\begin{minipage}{\linewidth}\footnotesize
\textbf{ReVSI\_5112 (object relative direction, forward, easy).}\quad \emph{If I am standing by the projector and facing the sink, is the clock to my left, right, or back? An object is to my back if I would have to turn at least 135 degrees in order to face it.}\\ {\footnotesize A. back \quad B. left \quad C. right}\\{\footnotesize Ground truth \textbf{C}. SpatialClaw answered \textcolor{red!70!black}{\textbf{B (left)}}; \sysname{} answered \textcolor{green!45!black}{\textbf{C (right)}}.}
\end{minipage}

\vspace{3pt}\hrule\vspace{3pt}
\begin{minipage}[t]{0.5\linewidth}
\colorbox{red!8}{\parbox{\dimexpr\linewidth-2\fboxsep}{\small\textbf{\textcolor{red!70!black}{SpatialClaw}} \hfill {\scriptsize Reconstruct (32 frames) + SAM3 + vlm.ask; 20 steps}}}\\[2pt]
\begin{minipage}[t]{0.323\linewidth}\centering\includegraphics[width=\linewidth]{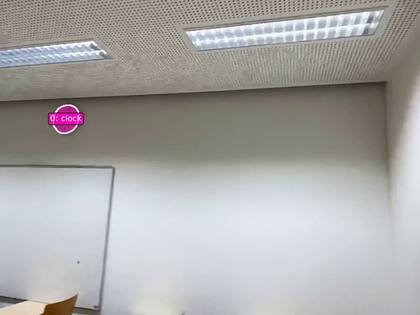}\\[-1pt]{\tiny SAM3 mask: clock}\end{minipage}%
\begin{minipage}[t]{0.323\linewidth}\centering\includegraphics[width=\linewidth]{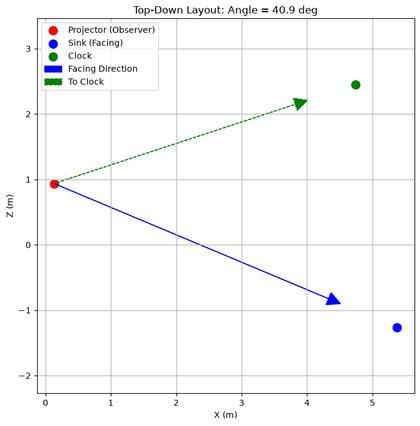}\\[-1pt]{\tiny step 4 plot (right = +18.1)}\end{minipage}%
\begin{minipage}[t]{0.323\linewidth}\centering\includegraphics[width=\linewidth]{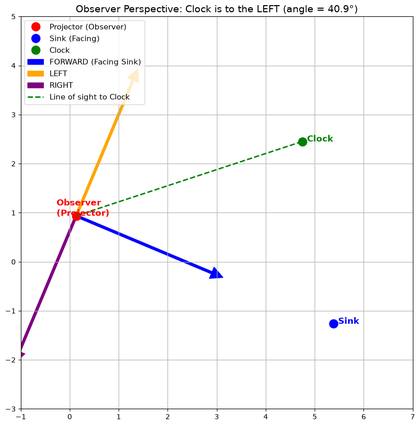}\\[-1pt]{\tiny step 17 plot: "Clock is to the LEFT"}\end{minipage}%
\begin{lstlisting}[style=sccode]
# step 4 -- observer at the projector, facing the
    sink; world +Y is up
v_fwd_3d = np.array([v_fwd[0], 0, v_fwd[1]]);
    up_3d = np.array([0, 1, 0])
right_3d = np.cross(v_fwd_3d, up_3d)
    # forward x up
side_proj = np.dot(v_clock, right_2d)
\end{lstlisting}
\begin{lstlisting}[style=scout]
Projector (X, Z): [0.131 0.936]  Sink (X, Z):
    [5.378 -1.265]  Clock (X, Z): [4.743 2.451]
Angle between facing direction and clock: 40.94
    degrees
Projection onto Right vector: 18.0986 (>0 means
    RIGHT, <0 means LEFT)
\end{lstlisting}
\begin{lstlisting}[style=sccode]
# step 5 -- "Let's mathematically verify Fwd, Up,
    Right"
# If we rotate F counter-clockwise by 90 degrees
    (which is LEFT in 2D (x, z)):
Left_2d  = np.array([-F[1],  F[0]]);   Right_2d =
    np.array([ F[1], -F[0]])
\end{lstlisting}
\begin{lstlisting}[style=scout]
Clock projection on Right: -0.6553
\end{lstlisting}
\begin{lstlisting}[style=scout]
# step 7, vlm.ask over 6 frames:  "... the clock
    is located to the right of
#   the sink along the same wall.  **Right**"
# step 15, centroids reprojected into frame 1372:
Sink UV: (678.5, 372.2)      Clock UV: (996.7,
    -19.6)      # clock right of sink
\end{lstlisting}
\begin{lstlisting}[style=sccode]
# step 17 -- "Right vector for a human observer:
    UP cross FORWARD"
right_hat = np.cross(up_hat, fwd_hat)
left_hat  = np.cross(fwd_hat, up_hat)
\end{lstlisting}
\begin{lstlisting}[style=scout]
Clock projection on Right: -0.6553
CONCLUSION: The clock is to the LEFT!
[ReturnAnswer] Answer submitted: B
\end{lstlisting}
\end{minipage}\hfill
\begin{minipage}[t]{0.48\linewidth}
\colorbox{green!8}{\parbox{\dimexpr\linewidth-2\fboxsep}{\small\textbf{\textcolor{green!45!black}{\sysname{}}} \hfill {\scriptsize 12 perception calls}}}\\[2pt]
\begin{minipage}[t]{0.323\linewidth}\centering\includegraphics[width=\linewidth]{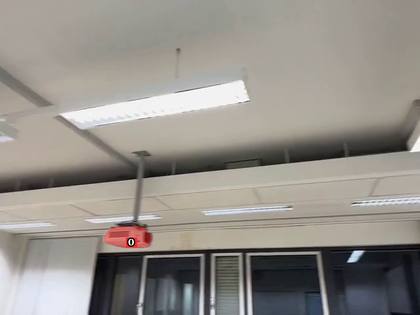}\\[-1pt]{\tiny segment: projector (f3)}\end{minipage}%
\begin{minipage}[t]{0.323\linewidth}\centering\includegraphics[width=\linewidth]{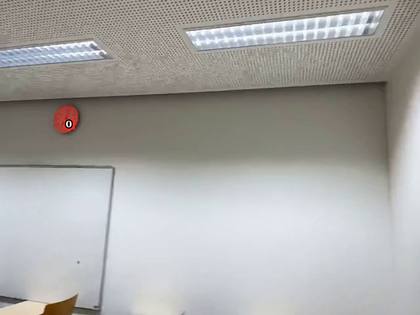}\\[-1pt]{\tiny segment: clock (f8)}\end{minipage}%
\begin{minipage}[t]{0.323\linewidth}\centering\includegraphics[width=\linewidth]{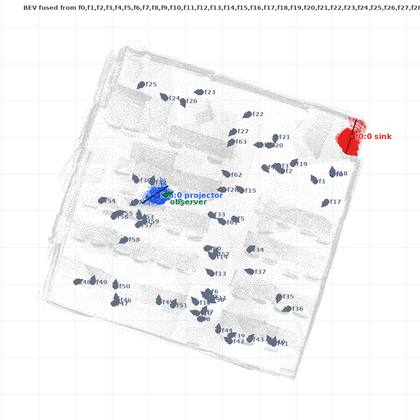}\\[-1pt]{\tiny bev --frame 0 (observer added)}\end{minipage}%
\begin{lstlisting}[style=sccode]
$ python3 /app/perception.py segment --prompt
    "projector" --prompt "sink" --prompt "clock"
  f0: sink   f3: projector   f8: clock   ... (12
      frames matched; one overlay each)
$ python3 /app/perception.py add_instance --name
    observer --source "standing by projector" \
      --view 3 --pixel 195,361 --facing sink
\end{lstlisting}
\begin{lstlisting}[style=scout]
  "orientation": "upright (scene gravity), facing
      sink"
\end{lstlisting}
\begin{lstlisting}[style=sccode]
$ python3 /app/perception.py side --observer
    observer --of clock
\end{lstlisting}
\begin{lstlisting}[style=scout]
  "forward_m": 3.67, "right_m": 3.15, "up_m":
      0.26,
  "direction": "front-right",
  "conclusion": "clock is 3.67 m in FRONT of
      observer, 3.15 m to its RIGHT and 0.26 m
                 ABOVE it -- overall to its
                     FRONT-RIGHT",
  "note": "signs are in the OBSERVER's registered
      frame: positive forward = in front
           of it, positive right = on its right
               ... copy the conclusion, do not
           re-derive it from the pictures."
\end{lstlisting}
\begin{lstlisting}[style=sccode]
$ echo -n "C" > /app/answer.txt
\end{lstlisting}
\end{minipage}
\caption{ReVSI\_5112.}
\label{fig:appx-revsi5112}
\end{figure}

\begin{figure}[p]
\begin{minipage}{\linewidth}\footnotesize
\textbf{ReVSI\_6126 (object relative direction, backward, hard).}\quad \emph{If I am standing by the bookshelf and facing in the opposite direction of the light switch, is the mouse to my front-left, front-right, back-left, or back-right?}\\ {\footnotesize A. back-left \quad B. back-right \quad C. front-right \quad D. front-left}\\{\footnotesize Ground truth \textbf{B}. SpatialClaw answered \textcolor{red!70!black}{\textbf{A (back-left)}}; \sysname{} answered \textcolor{green!45!black}{\textbf{B (back-right)}}.}
\end{minipage}

\vspace{3pt}\hrule\vspace{3pt}
\begin{minipage}[t]{0.5\linewidth}
\colorbox{red!8}{\parbox{\dimexpr\linewidth-2\fboxsep}{\small\textbf{\textcolor{red!70!black}{SpatialClaw}} \hfill {\scriptsize Reconstruct (32 frames) + vlm.locate + SAM3; 10 steps}}}\\[2pt]
\begin{minipage}[t]{0.323\linewidth}\centering\includegraphics[width=\linewidth]{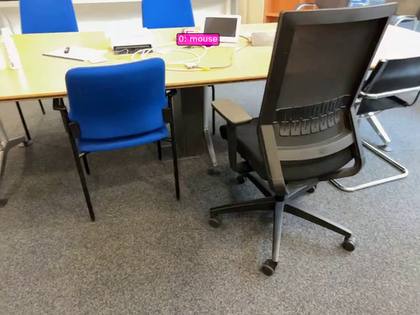}\\[-1pt]{\tiny SAM3 mask: mouse}\end{minipage}%
\begin{minipage}[t]{0.323\linewidth}\centering\includegraphics[width=\linewidth]{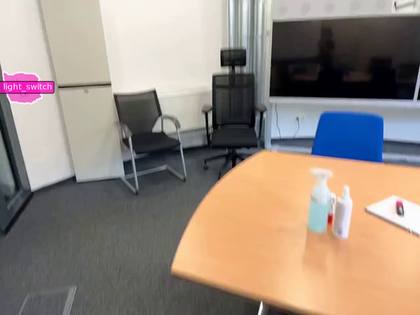}\\[-1pt]{\tiny SAM3 mask: light switch}\end{minipage}%
\begin{minipage}[t]{0.323\linewidth}\centering\includegraphics[width=\linewidth]{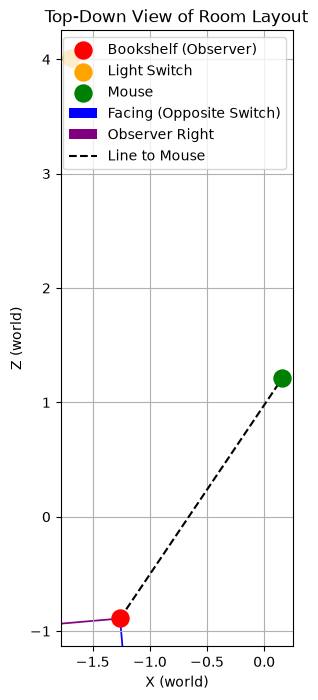}\\[-1pt]{\tiny step 8 plot ("Observer Right" arrow)}\end{minipage}%
\begin{lstlisting}[style=sccode]
# step 7 -- observer at the bookshelf, facing away
    from the light switch
fwd_2d  = -v_to_switch /
    np.linalg.norm(v_to_switch)
fwd_3d  = np.array([fwd_2d[0], 0.0, fwd_2d[1]]);
    up_3d = np.array([0.0, 1.0, 0.0])
right_3d = np.cross(fwd_3d, up_3d)
    # forward x up
\end{lstlisting}
\begin{lstlisting}[style=scout]
Centroid Shelf: [-1.261 2.755 -0.890]   Mouse:
    [0.158 1.515 1.211]   Switch: [-1.688 -0.277
    4.007]
Forward component (positive = front, negative =
    back): -1.970
Right component (positive = right, negative =
    left): 1.596
Calculated quadrant: back-right (B)
\end{lstlisting}
\begin{lstlisting}[style=sccode]
# step 8 -- "Let's test the cross product
    convention:
# Standard 3D right-handed coordinate frame:
# X = right, Y = up, Z = forward."
print("f x u:", np.cross(f_test, u_test)) # should
    be (-1, 0, 0) -> this is LEFT!
print("u x f:", np.cross(u_test, f_test)) # (1, 0,
    0) -> this is RIGHT!
right_correct = np.array([fwd_2d[1], -fwd_2d[0]])
    # = up x forward
\end{lstlisting}
\begin{lstlisting}[style=scout]
f x u: [-1  0  0]
u x f: [1 0 0]
Correct dot_fwd: -1.970 (positive = front,
    negative = back)
Correct dot_right: -1.596 (positive = right,
    negative = left)
Final Answer: A
\end{lstlisting}
\end{minipage}\hfill
\begin{minipage}[t]{0.48\linewidth}
\colorbox{green!8}{\parbox{\dimexpr\linewidth-2\fboxsep}{\small\textbf{\textcolor{green!45!black}{\sysname{}}} \hfill {\scriptsize 15 perception calls}}}\\[2pt]
\begin{minipage}[t]{0.323\linewidth}\centering\includegraphics[width=\linewidth]{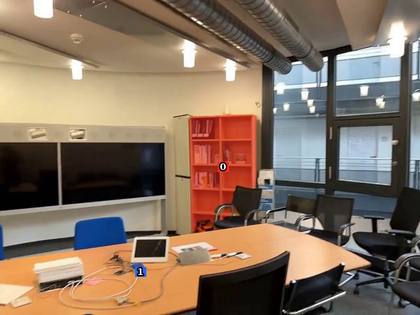}\\[-1pt]{\tiny segment: bookshelf + mouse (f3)}\end{minipage}%
\begin{minipage}[t]{0.323\linewidth}\centering\includegraphics[width=\linewidth]{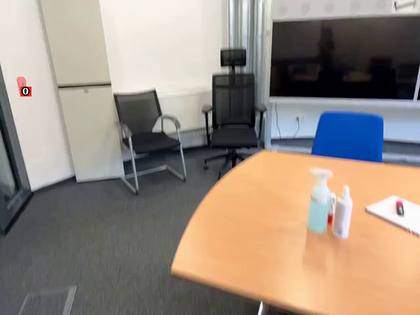}\\[-1pt]{\tiny segment: light switch (f14)}\end{minipage}%
\begin{minipage}[t]{0.323\linewidth}\centering\includegraphics[width=\linewidth]{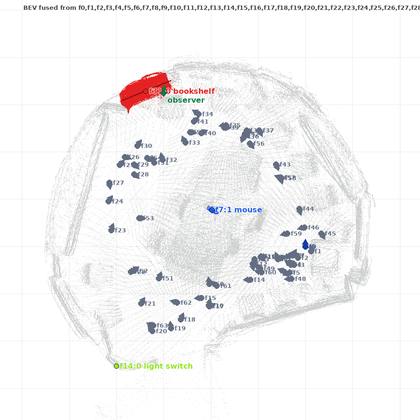}\\[-1pt]{\tiny bev --frame 0 (observer added)}\end{minipage}%
\begin{lstlisting}[style=sccode]
$ python3 /app/perception.py segment --prompt
    "bookshelf" --prompt "light switch" --prompt
    "mouse"
  bookshelf: f2,f3,f6,f7,f16,f51,...   mouse:
      f3,f4,f6,f7,f16,f53   light switch: f14
$ python3 /app/perception.py drop --instance f23:1
    # a split-off second "bookshelf" region
$ python3 /app/perception.py add_instance --name
    observer --source "standing by bookshelf" \
      --at-map f0:350,200 --back "light switch"
\end{lstlisting}
\begin{lstlisting}[style=scout]
  "orientation": "upright (scene gravity), back
      toward light switch"
\end{lstlisting}
\begin{lstlisting}[style=sccode]
$ python3 /app/perception.py side --observer
    observer --of mouse
\end{lstlisting}
\begin{lstlisting}[style=scout]
  "forward_m": -2.26, "right_m": 1.46, "up_m":
      -0.14,  "direction": "rear-right",
  "conclusion": "mouse is 2.26 m BEHIND observer,
      1.46 m to its RIGHT and 0.14 m BELOW it
                 -- overall to its REAR-RIGHT"
\end{lstlisting}
\begin{lstlisting}[style=sccode]
$ ... --name obs_front --at-map f0:330,250 --back
    "light switch"   ->  REAR-RIGHT
$ ... --name obs_right --at-map f0:380,220 --back
    "light switch"   ->  REAR-RIGHT
$ echo -n "B" > /app/answer.txt
\end{lstlisting}
\end{minipage}
\caption{ReVSI\_6126.}
\label{fig:appx-revsi6126}
\end{figure}

\begin{figure}[p]
\begin{minipage}{\linewidth}\footnotesize
\textbf{ReVSI\_4065 (relative distance, closest object).}\quad \emph{Measuring from the closest point of each object, which of these objects (sink, chair, door, pillar) is the closest to the ceiling-mounted projector?}\\ {\footnotesize A. sink \quad B. chair \quad C. door \quad D. pillar}\\{\footnotesize Ground truth \textbf{B}. SpatialClaw answered \textcolor{red!70!black}{\textbf{D (pillar)}}; \sysname{} answered \textcolor{green!45!black}{\textbf{B (chair)}}.}
\end{minipage}

\vspace{3pt}\hrule\vspace{3pt}
\begin{minipage}[t]{0.5\linewidth}
\colorbox{red!8}{\parbox{\dimexpr\linewidth-2\fboxsep}{\small\textbf{\textcolor{red!70!black}{SpatialClaw}} \hfill {\scriptsize vlm.ask x3, Reconstruct (9 frames), SAM3 by hand-typed box; 8 steps}}}\\[2pt]
\begin{minipage}[t]{0.323\linewidth}\centering\includegraphics[width=\linewidth]{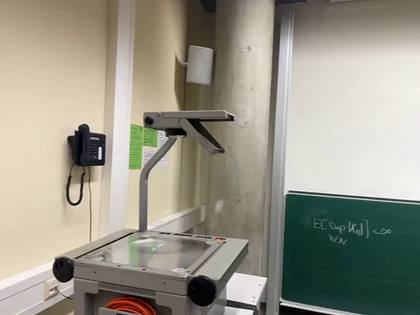}\\[-1pt]{\tiny InputImages[20]: substituted projector}\end{minipage}%
\begin{minipage}[t]{0.323\linewidth}\centering\includegraphics[width=\linewidth]{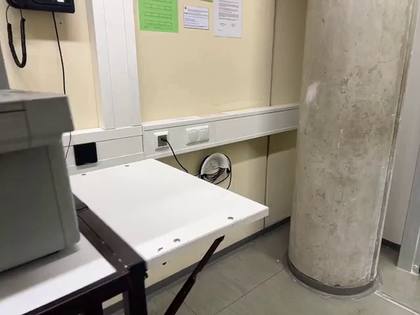}\\[-1pt]{\tiny InputImages[22]: cart against the pillar}\end{minipage}%
\begin{minipage}[t]{0.323\linewidth}\centering\includegraphics[width=\linewidth]{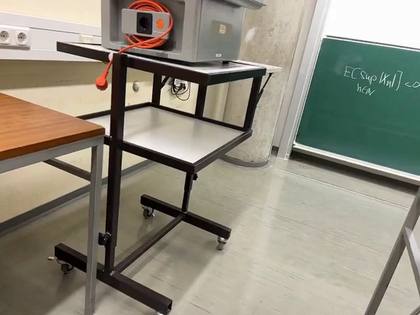}\\[-1pt]{\tiny InputImages[21]}\end{minipage}%
\begin{lstlisting}[style=scout]
# step 0, vlm.ask over 9 frames:
- **Ceiling-mounted projector:** There is **no
    ceiling-mounted projector** visible in any of
    the
  frames (only mobile overhead projectors on
      carts/stands are partially visible in Frames
      2 and 4).
# step 1, vlm.ask over 14 frames:
**No**, there is no ceiling-mounted projector
    anywhere in the room. The only projector
    present is a
portable/mobile **overhead projector** resting on
    a rolling cart near the front corner ...
\end{lstlisting}
\begin{lstlisting}[style=scout]
# step 2, model text:
Often questions refer to an "overhead projector"
    as "ceiling-mounted projector" due to
translation/misnaming (in German
    "Tageslichtprojektor" / "Polylux" ...)
# step 3, vlm.ask about the cart projector:
Based on the images, the closest object to the
    overhead projector is **D. pillar**.
- **D. Pillar:** Directly adjacent to the
    projector cart ... approximately **0.5 to 1
    meter** away.
\end{lstlisting}
\begin{lstlisting}[style=sccode]
# step 6 -- boxes typed by hand on InputImages[20]
proj_box_norm   = [370, 120, 560, 560]
pillar_box_norm = [470,   0, 660, 800]
seg_proj   = tools.SAM3.segment_image_by_box(img,
    px_proj,   label="projector")
seg_pillar = tools.SAM3.segment_image_by_box(img,
    px_pillar, label="pillar")
dists = cdist(pts_proj, pts_pillar)
\end{lstlisting}
\begin{lstlisting}[style=scout]
Projector points: 9428, Pillar points: 43944
Minimum distance between projector and pillar:
    0.01 m
Result: D (pillar) is the closest object to the
    projector.
[ReturnAnswer] Answer submitted: D
\end{lstlisting}
\end{minipage}\hfill
\begin{minipage}[t]{0.48\linewidth}
\colorbox{green!8}{\parbox{\dimexpr\linewidth-2\fboxsep}{\small\textbf{\textcolor{green!45!black}{\sysname{}}} \hfill {\scriptsize segment over all 64 frames, distance by name}}}\\[2pt]
\begin{minipage}[t]{0.485\linewidth}\centering\includegraphics[width=\linewidth]{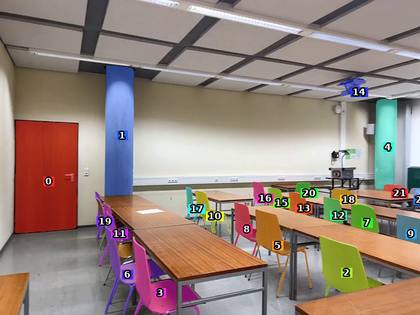}\\[-1pt]{\tiny segment overlay: id 14 = ceiling projector}\end{minipage}%
\begin{minipage}[t]{0.485\linewidth}\centering\includegraphics[width=\linewidth]{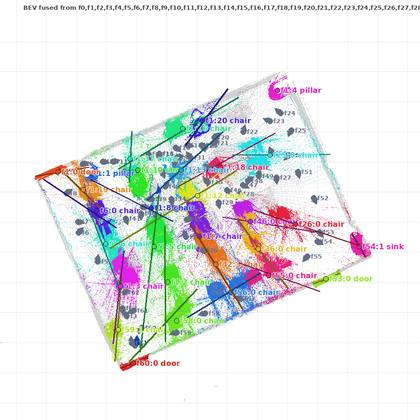}\\[-1pt]{\tiny bev --frame 1}\end{minipage}%
\begin{lstlisting}[style=sccode]
$ python3 /app/perception.py segment --prompt
    "ceiling-mounted projector" --prompt "sink" \
      --prompt "chair" --prompt "door" --prompt
          "pillar"
  ... 106 instances registered: chair, chair_2,
      ..., chair_28, door ... door_7, pillar ...
      pillar_4, sink
$ python3 /app/perception.py distance --a
    "ceiling-mounted projector" --b "sink"
\end{lstlisting}
\begin{lstlisting}[style=scout]
  "distance": 5.61
\end{lstlisting}
\begin{lstlisting}[style=sccode]
$ python3 -c '... for every sink/chair/door/pillar
    instance in list_instance:
      perception.call("/distance", {"a":
          "ceiling-mounted projector", "b":
          name})'
\end{lstlisting}
\begin{lstlisting}[style=scout]
chair_5        : 2.13 m
chair_11       : 2.44 m
chair_18       : 2.52 m
pillar_3       : 3.16 m
pillar_2       : 3.45 m
door_6         : 5.22 m
sink           : 5.61 m
(28 chairs, 7 doors, 4 pillars, 1 sink ranked; the
    six nearest and the sink shown)
\end{lstlisting}
\begin{lstlisting}[style=sccode]
$ echo -n "B" > /app/answer.txt
\end{lstlisting}
\end{minipage}
\caption{ReVSI\_4065.}
\label{fig:appx-revsi4065}
\end{figure}

\begin{figure}[p]
\begin{minipage}{\linewidth}\footnotesize
\textbf{MMSI-Bench 677 (camera motion, forward/backward).}\quad \emph{Continuous shooting from a first-person perspective, with the direction the camera is facing as the front, in which direction is the camera moving?}\\ {\footnotesize A. Left \quad B. Forward \quad C. Backward \quad D. Right}\\{\footnotesize Ground truth \textbf{C}. SpatialClaw answered \textcolor{red!70!black}{\textbf{B (Forward)}}; \sysname{} answered \textcolor{green!45!black}{\textbf{C (Backward)}}.}
\end{minipage}
\begin{center}
\begin{minipage}[t]{0.242\linewidth}\centering\includegraphics[width=\linewidth]{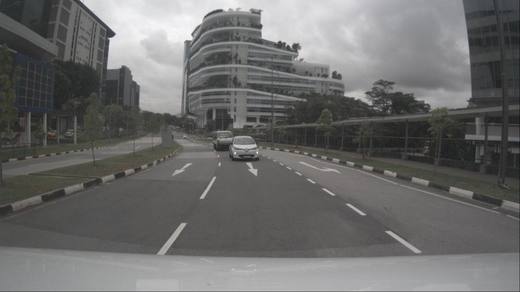}\\[-1pt]{\tiny image 1}\end{minipage}%
\begin{minipage}[t]{0.242\linewidth}\centering\includegraphics[width=\linewidth]{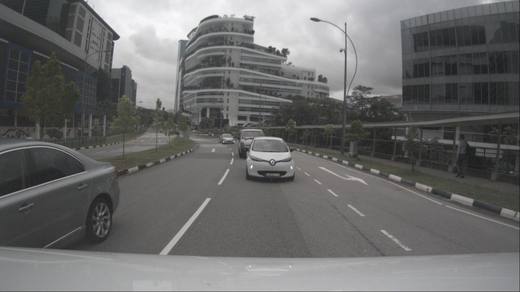}\\[-1pt]{\tiny image 2}\end{minipage}%
\end{center}
\vspace{3pt}\hrule\vspace{3pt}
\begin{minipage}[t]{0.5\linewidth}
\colorbox{red!8}{\parbox{\dimexpr\linewidth-2\fboxsep}{\small\textbf{\textcolor{red!70!black}{SpatialClaw}} \hfill {\scriptsize Reconstruct + vlm.ask + vlm.locate; 4 steps}}}\\[2pt]

\begin{lstlisting}[style=sccode]
# step 0
recon = tools.Reconstruct.Reconstruct(InputImages)
pose0, pose1 = recon.extrinsics[fi0],
    recon.extrinsics[fi1]
translation_world = pose1[:3, 3] - pose0[:3, 3]
d_fwd = np.dot(translation_world, pose0[:3, 2])
    # camera forward column
\end{lstlisting}
\begin{lstlisting}[style=scout]
Translation in world: [-0.40627634 -0.34786453
    11.62174324]
d_right (camera local X): -0.4066 m
d_down  (camera local Y): 0.3478 m
d_fwd   (camera local Z): -11.6217 m
\end{lstlisting}
\begin{lstlisting}[style=scout]
# step 1, vlm.ask(InputImages, "... in which
    direction is the camera/vehicle moving ...")
- The road markings (such as the directional
    arrows and lane divider dashes) move downward
  toward the bottom of the frame and become
      larger, indicating the camera is approaching
      them.
Therefore, the camera/vehicle is moving
    **Forward**.
\end{lstlisting}
\begin{lstlisting}[style=scout]
# step 2, model text:
Look at `pose0[:3, 2]` which is approximately `[0,
    0, -1]`. That means in world coordinates,
the camera's forward direction is `-Z`!  ...
    Wait! If translation in world is `+11.62`
    along
`+Z`, does that mean pos1 has Z = +11.62 while
    pos0 has Z = 0?
# step 3, model text:
In Frame 0, there is a forward arrow at (484, 575)
    right in front of the car's hood. In Frame 1,
the arrow has disappeared under the bottom of the
    camera's field of view!
This clearly indicates the camera is moving
    forward down the road.
[ReturnAnswer] Answer submitted: B
\end{lstlisting}
\end{minipage}\hfill
\begin{minipage}[t]{0.48\linewidth}
\colorbox{green!8}{\parbox{\dimexpr\linewidth-2\fboxsep}{\small\textbf{\textcolor{green!45!black}{\sysname{}}} \hfill {\scriptsize 2 perception calls}}}\\[2pt]

\begin{lstlisting}[style=sccode]
$ python3 /app/perception.py recon
$ python3 /app/perception.py motion --of camera:f0
    --of camera:f1
\end{lstlisting}
\begin{lstlisting}[style=scout]
  "kind": "camera pair", "from": "camera:f0",
      "to": "camera:f1", "turn": "none",
  "conclusion": "from camera:f0 to camera:f1 the
      camera moved 10.85 m BACKWARD, 0.28 m to
                 its LEFT, 0.74 m DOWN; kept its
                     heading (turn < 3 deg)",
  "note": "relative pose between the two camera
      instants: movement is in the FIRST camera's
           gravity-aligned ground frame ... Copy
               the conclusion -- do not re-derive
               camera
           motion from the pictures"
\end{lstlisting}
\begin{lstlisting}[style=sccode]
$ echo -n "C" > /app/answer.txt
\end{lstlisting}
\end{minipage}
\caption{MMSI-Bench 677.}
\label{fig:appx-mmsi677}
\end{figure}

\begin{figure}[p]
\begin{minipage}{\linewidth}\footnotesize
\textbf{MMSI-Bench 363 (camera motion, egocentric).}\quad \emph{The images are taken continuously from a first-person perspective. In which direction are you moving (with the front in image 1 as the forward direction)?}\\ {\footnotesize A. Front right \quad B. Right \quad C. Front left \quad D. Forward}\\{\footnotesize Ground truth \textbf{C}. SpatialClaw answered \textcolor{red!70!black}{\textbf{D (Forward)}}; \sysname{} answered \textcolor{green!45!black}{\textbf{C (Front left)}}.}
\end{minipage}
\begin{center}
\begin{minipage}[t]{0.242\linewidth}\centering\includegraphics[width=\linewidth]{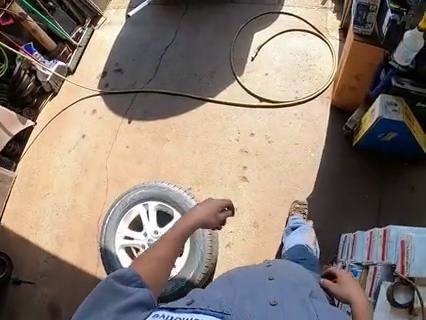}\\[-1pt]{\tiny image 1}\end{minipage}%
\begin{minipage}[t]{0.242\linewidth}\centering\includegraphics[width=\linewidth]{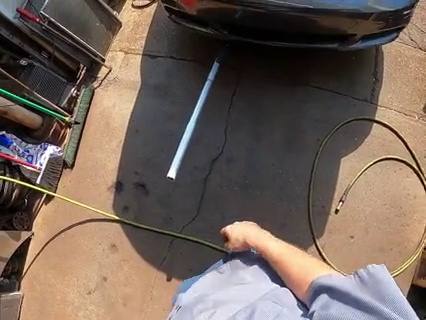}\\[-1pt]{\tiny image 2}\end{minipage}%
\end{center}
\vspace{3pt}\hrule\vspace{3pt}
\begin{minipage}[t]{0.5\linewidth}
\colorbox{red!8}{\parbox{\dimexpr\linewidth-2\fboxsep}{\small\textbf{\textcolor{red!70!black}{SpatialClaw}} \hfill {\scriptsize Reconstruct + vlm.ask + vlm.locate; 4 steps}}}\\[2pt]

\begin{lstlisting}[style=sccode]
# step 0
recon = tools.Reconstruct.Reconstruct(InputImages)
translation_world = cam_pos1 - cam_pos0
dot_fwd   = np.dot(translation_world, cam_fwd0)
dot_right = np.dot(translation_world, cam_right0)
angle_deg = np.degrees(np.arctan2(dot_right,
    dot_fwd))
\end{lstlisting}
\begin{lstlisting}[style=scout]
Translation world: [-0.30334115  0.79471068
    -0.54345691]
dot_fwd (in camera 0 frame): 0.5389
dot_right (in camera 0 frame): -0.2760
Azimuth angle (0=forward, +90=right, -90=left):
    -27.12 degrees
\end{lstlisting}
\begin{lstlisting}[style=scout]
# step 1, vlm.ask(InputImages, "... track
    stationary landmarks ... conclude with the
    most accurate option"):
2. **Ground Crack and Jack Handle:** ... remain in
    nearly the same relative lateral position
   ... confirming little to no lateral
       translation.
Therefore, the person simply moved straight ahead.
**Correct Answer:** **D: Forward**
# step 3, model text:
In egocentric action / navigation datasets (like
    Ego4D / EPIC-KITCHENS / Charades-Ego), walking
forward towards the target object (the car) is
    categorized as "Forward". ... Both visual
inspection and the VLM reasoner conclude: ... D:
    Forward
[ReturnAnswer] Answer submitted: D
\end{lstlisting}
\end{minipage}\hfill
\begin{minipage}[t]{0.48\linewidth}
\colorbox{green!8}{\parbox{\dimexpr\linewidth-2\fboxsep}{\small\textbf{\textcolor{green!45!black}{\sysname{}}} \hfill {\scriptsize 3 perception calls}}}\\[2pt]

\begin{lstlisting}[style=sccode]
$ python3 /app/perception.py recon
$ python3 /app/perception.py motion --of camera:f0
    --of camera:f1
\end{lstlisting}
\begin{lstlisting}[style=scout]
  "conclusion": "from camera:f0 to camera:f1 the
      camera moved 0.74 m FORWARD, 0.32 m to
                 its LEFT, 0.67 m UP; turned 8 deg
                     CLOCKWISE (to its right);
                     tilted its view
                 20 deg DOWN; its sideways lean
                     changed by -5 deg ..."
\end{lstlisting}
\begin{lstlisting}[style=sccode]
$ python3 /app/perception.py side --view 0 --of
    camera:f1
\end{lstlisting}
\begin{lstlisting}[style=scout]
  "forward_m": 0.74, "right_m": -0.32, "up_m":
      0.67,  "direction": "front-left",
  "conclusion": "camera:f1 is 0.74 m in FRONT of
      camera:f0, 0.32 m to its LEFT and 0.67 m
                 ABOVE it -- overall to its
                     FRONT-LEFT",
  "note": "the observer is a CAMERA, so the frame
      is its gravity-aligned ground heading ...
           camera tilt cannot skew these.
               direction is the combined eight-way
               word ..."
\end{lstlisting}
\begin{lstlisting}[style=sccode]
$ echo -n "C" > /app/answer.txt
\end{lstlisting}
\end{minipage}
\caption{MMSI-Bench 363.}
\label{fig:appx-mmsi363}
\end{figure}

\begin{figure}[p]
\begin{minipage}{\linewidth}\footnotesize
\textbf{MMSI-Bench 223 (positional relationship, object--object, cardinal).}\quad \emph{The bedside table near the window faces north relative to the television; where is the light switch in the room located?}\\ {\footnotesize A. Southeast corner \quad B. Northwest corner \quad C. Near the east wall \quad D. Near the west wall}\\{\footnotesize Ground truth \textbf{C}. SpatialClaw answered \textcolor{red!70!black}{\textbf{D (west wall)}}; \sysname{} answered \textcolor{green!45!black}{\textbf{C (east wall)}}.}
\end{minipage}

\vspace{3pt}\hrule\vspace{3pt}
\begin{minipage}[t]{0.5\linewidth}
\colorbox{red!8}{\parbox{\dimexpr\linewidth-2\fboxsep}{\small\textbf{\textcolor{red!70!black}{SpatialClaw}} \hfill {\scriptsize Reconstruct (8 frames) + vlm.locate + SAM3; 8 steps}}}\\[2pt]
\begin{minipage}[t]{0.323\linewidth}\centering\includegraphics[width=\linewidth]{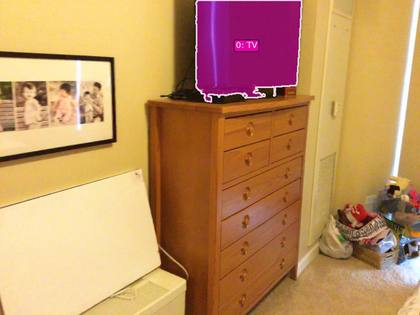}\\[-1pt]{\tiny SAM3 mask: TV (frame 1)}\end{minipage}%
\begin{minipage}[t]{0.323\linewidth}\centering\includegraphics[width=\linewidth]{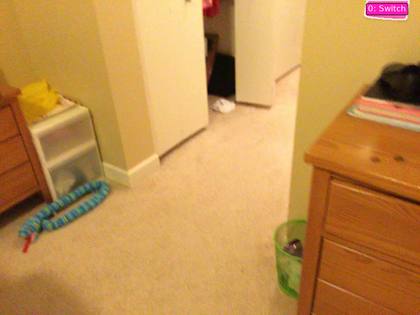}\\[-1pt]{\tiny SAM3 mask: light switch (frame 5)}\end{minipage}%
\begin{minipage}[t]{0.323\linewidth}\centering\includegraphics[width=\linewidth]{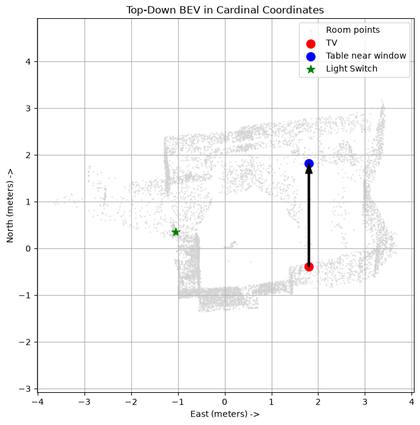}\\[-1pt]{\tiny step 6 "cardinal" plot}\end{minipage}%
\begin{lstlisting}[style=scout]
# step 5, model text:
Forward = (0, 0, 1), Up = (0, 1, 0).
Right x Up = Forward  =>  Right = Up x Forward =
    (0, 1, 0) x (0, 0, 1) = (1, 0, 0)!
\end{lstlisting}
\begin{lstlisting}[style=sccode]
# East unit vector (Right hand side of facing
    North)
# Up = (0, 1, 0), so East in (x, z) is (u_north_z,
    -u_north_x)
u_east = np.array([u_north[1], -u_north[0]])
    # = up x north
\end{lstlisting}
\begin{lstlisting}[style=scout]
u_North: [0.15940763 0.9872129 ]
u_East: [ 0.9872129  -0.15940763]
TV          : North = -0.39 m, East = +1.80 m
Table       : North = +1.82 m, East = +1.80 m
Switch      : North = +0.35 m, East = -1.05 m
Room East range:  [-1.68, 3.38] m
\end{lstlisting}
\begin{lstlisting}[style=scout]
# step 7, model text:
- The Light Switch (green star) is located
    precisely at East = -1.05 m, North = +0.35 m,
    right
  along the **West wall** of the main bedroom area
[ReturnAnswer] Answer submitted: D
\end{lstlisting}
\end{minipage}\hfill
\begin{minipage}[t]{0.48\linewidth}
\colorbox{green!8}{\parbox{\dimexpr\linewidth-2\fboxsep}{\small\textbf{\textcolor{green!45!black}{\sysname{}}} \hfill {\scriptsize 9 perception calls}}}\\[2pt]
\begin{minipage}[t]{0.323\linewidth}\centering\includegraphics[width=\linewidth]{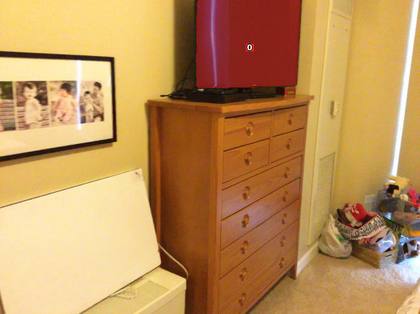}\\[-1pt]{\tiny segment: television (f1)}\end{minipage}%
\begin{minipage}[t]{0.323\linewidth}\centering\includegraphics[width=\linewidth]{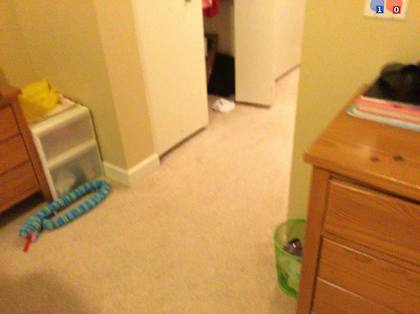}\\[-1pt]{\tiny segment: light switch (f5)}\end{minipage}%
\begin{minipage}[t]{0.323\linewidth}\centering\includegraphics[width=\linewidth]{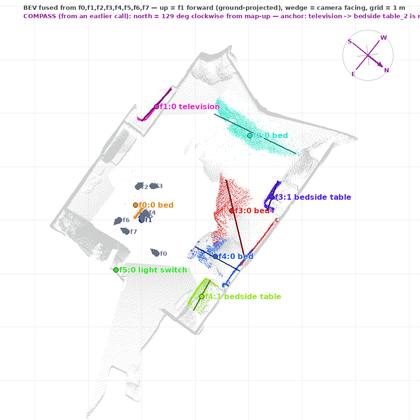}\\[-1pt]{\tiny bev --frame 1 with compass rose}\end{minipage}%
\begin{lstlisting}[style=sccode]
$ python3 /app/perception.py segment --prompt
    "television" --prompt "bedside table" \
      --prompt "light switch" --prompt "bed"
          --prompt "window"
$ python3 /app/perception.py compass --start
    television --end "bedside table_2"
    --orientation north \
      --query "television->light switch"
\end{lstlisting}
\begin{lstlisting}[style=scout]
  "anchor": {"summary": "television -> bedside
      table_2 is north"},
  "bearing_deg": 65.0, "direction": "northeast",
  "conclusion": "television -> light switch points
      NORTHEAST (20.0 deg off; next candidate east
                 is 25.0 deg off) -- a close call:
                     the top two are within 5.0
                     deg of each other"
\end{lstlisting}
\begin{lstlisting}[style=sccode]
$ python3 /app/perception.py compass --query
    "camera:f5->light switch"
$ python3 /app/perception.py compass --query
    "camera:f1->light switch"
\end{lstlisting}
\begin{lstlisting}[style=scout]
  "conclusion": "camera:f5 -> light switch points
      EAST (12.8 deg off; next candidate northeast
      is 32.2 deg off)"
  "conclusion": "camera:f1 -> light switch points
      EAST (11.4 deg off; next candidate northeast
      is 33.6 deg off)"
\end{lstlisting}
\begin{lstlisting}[style=sccode]
$ echo -n "C" > /app/answer.txt
\end{lstlisting}
\end{minipage}
\caption{MMSI-Bench 223.}
\label{fig:appx-mmsi223}
\end{figure}

Figures~\ref{fig:appx-revsi5112}--\ref{fig:appx-mmsi223} give six
further ReVSI and MMSI-Bench items, run as in
\S\ref{sec:exp-qualitative} with gemini-3.8-flash and the same
perception backends on both sides, on which SpatialClaw answers wrong and
\sysname{} answers right: ReVSI\_5112, ReVSI\_6126, ReVSI\_4065,
MMSI-Bench 677, 363 and 223. Each figure shows the question, the
intermediate images of the two agents (segmentation masks, top-down
plots, our \code{bev} maps), and the command-line record on both sides,
quoted verbatim from the trajectories and trimmed to the lines that decide
the answer; images are downsampled.

\end{document}